\documentclass{article}

\usepackage[final]{neurips_2024}

\usepackage[utf8]{inputenc} 
\usepackage[T1]{fontenc}    
\usepackage{hyperref}       
\usepackage{url}            
\usepackage{booktabs}       
\usepackage{amsfonts}       
\usepackage{nicefrac}       
\usepackage{microtype}      
\usepackage{xcolor}         
\usepackage{graphicx} 
\usepackage{color}
\usepackage{amsmath, amssymb,amsfonts,amsthm}
\usepackage{dsfont}
\usepackage{caption}
\usepackage{comment}
\usepackage{makecell}
\usepackage{enumitem}
\usepackage{subcaption}
\usepackage{wrapfig}

\newcommand{\dejavu}{\emph{déjà vu }}
\newcommand{\Dejavu}{\emph{Déjà vu }}

\newtheorem{definition}{Definition}

\newcommand{\calD}{\mathcal{D}}
\newcommand{\argmax}{\mathrm{argmax}}

\title{Measuring \Dejavu Memorization Efficiently}

\author{Narine Kokhlikyan \\
FAIR at Meta \\
\And
Bargav Jayaraman \\
FAIR at Meta \\
\And
Florian Bordes \\
FAIR at Meta \\
\AND Chuan Guo \\
FAIR at Meta \\
\And Kamalika Chaudhuri \\
FAIR at Meta}

\begin{document}
\maketitle

\begin{abstract}
    Recent research has shown that representation learning models may accidentally memorize their training data. For example, the \dejavu method shows that for certain representation learning models and training images, it is sometimes possible to correctly predict the foreground label given only the representation of the background -- better than through dataset-level correlations. 
    However, their measurement method requires training two models -- one to estimate dataset-level correlations and the other to estimate memorization. This multiple model setup becomes infeasible for large open-source models.
     In this work, we propose alternative simple methods to estimate dataset-level correlations, and show that these can be used to approximate an off-the-shelf model's memorization ability without any retraining. This enables, for the first time, the measurement of memorization in pre-trained open-source image representation and vision-language representation models. Our results show that different ways of measuring memorization yield very similar aggregate results. We also find that open-source models typically have lower aggregate memorization than similar models trained on a subset of the data. The code is available both for \href{https://github.com/facebookresearch/DejaVuOSS}{\color{blue} vision} and \href{https://github.com/facebookresearch/VLMDejaVu}{\color{blue} vision language models}. 
\end{abstract}

\section{Introduction}

Representation learning has emerged as one of the major tasks in computer vision. The goal in representation learning is to learn a model that produces semantically meaningful representations, where images or image-text pairs that are close in meaning occur close together in representation space. These learned representations can then be used in numerous downstream applications such as semantic segmentation~\citep{kirillov2023segany}, image generation~\citep{rombach2022high} and multi-modal LLMs~\citep{liu2024visual}.
A natural question that arises is whether these representation learning models memorize their training data and to what extent. Excessive memorization may call the generalization abilities of the models into question. In addition, studying memorization helps us understand potential privacy risks associated with these models, Thus, there is a need to develop a way to measure if and to what extent memorization is taking place.

\begin{figure}[t]
    \centering
    \includegraphics[width=\textwidth]{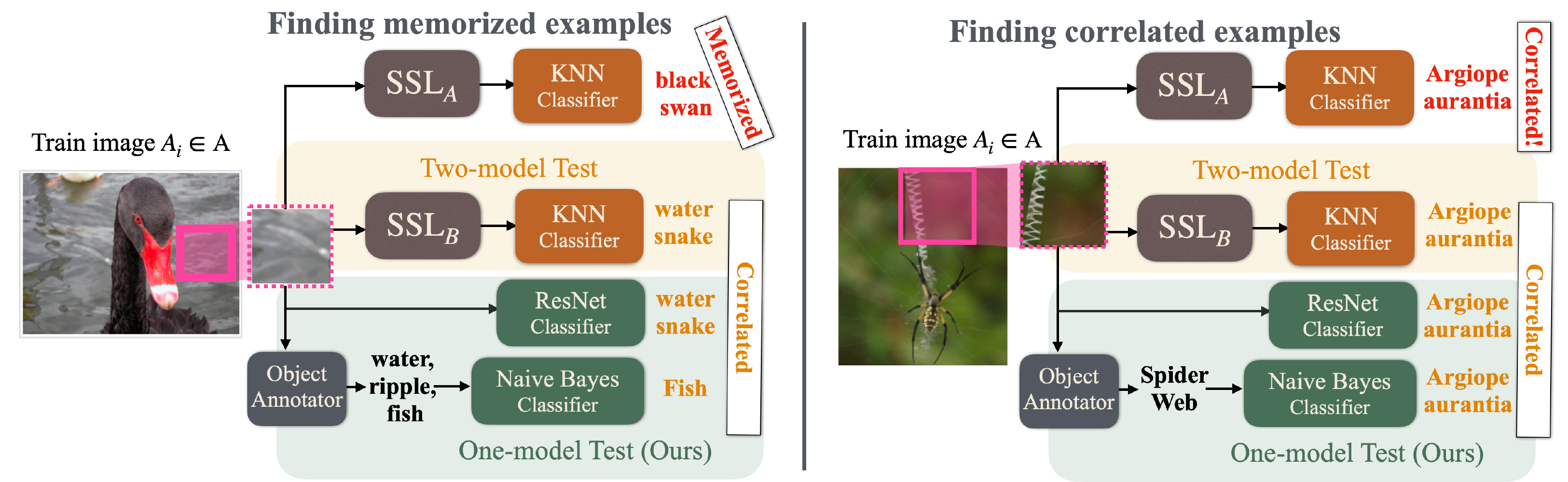}
    \caption{Illustration of our one-model \dejavu test for image representation learning. The task is to predict the foreground object given a background crop. The original \dejavu test~\citep{Dejavu} trains two models $\text{SSL}_A$ and $\text{SSL}_B$ on disjoint splits of the training set, and uses $\text{SSL}_B$ to quantify the degree of dataset-level correlation between the foreground and background crop. Our one-model test replaces $\text{SSL}_B$ with a classifier that directly predicts the foreground given background crop, and we show that both ResNet50 network and Naive Bayes classifier work well for this purpose.}
    \label{fig:intro_figure}
    \vspace{-3ex}
\end{figure}

Since learned representations are usually abstract and hard to interpret, memorization measurement for representation learning models requires careful design.
Currently, a standard way of doing this is the \dejavu method, which designs a causal task of predicting parts of the training sample given another disjoint part, and uses performance on this task to determine if the model memorizes. For example, \citet{Dejavu} designed the task of predicting the foreground object given the background crop of a training image, as shown in \autoref{fig:intro_figure} (orange block). Achieving a high performance on this task indicates two possibilities: \emph{(i)} if the model has memorized the association of the background crop with the foreground object for a specific training sample, or \emph{(ii)} if the model has learned the dataset-level correlation between the background crop and a given foreground object. To rule out the second possibility, \citet{Dejavu} opted for a two-model approach, training two separate models on disjoint parts of the training set and using the gap in performance between the two models as indication of memorization. This approach has been extended by \citet{jayaraman2024deja} to measure memorization of vision-language models such as CLIP~\citep{radford2021learning}.

Despite the success of the \dejavu method in defining and measuring memorization, scaling this approach to state-of-the-art representation learning models is challenging. First, the two-model approach requires the model trainer to split the training set into disjoint halves, which severely constrains valuable training data. Moreover, even if data is abundant, training the second model on internet-scale datasets is computationally expensive. Due to these limitations, the \dejavu test cannot be used to measure memorization of pre-trained models out-of-the-box.

In this work, we provide simple alternative ways of quantifying dataset-level correlations and show that they suffice for the purpose of measuring \dejavu memorization. Specifically, for image representation learning, we propose two alternative ways to derive reference models to predict the foreground label from a background crop: training an image classification network directly, and using a Naive Bayes classifier on top of a pre-trained object detection model. We then leverage these reference models to define a \emph{one-model} \dejavu test---a memorization test for representation learning models that only requires training a simpler reference model \emph{once per dataset}. \autoref{fig:intro_figure} (green block) gives an illustration of our proposed one-model \dejavu test. We also propose a variant of the method for vision-language models by leveraging a pre-trained text embedding model.

We validate our proposed methods by comparing them to the two-model test on ImageNet-trained image representation learning models, as well as CLIP models trained on a privately licensed image-caption pair dataset. We find that the one-model test can successfully identify memorized examples and obtain similar population-level memorization scores as the two-model test. We then apply the one-model \dejavu test on pre-trained open-source models and provide for the first time a principled memorization measurement on these models. Our results reveal that open-source models have significantly lower memorization rates than similar models trained on a smaller subset of data. We conclude that our one-model test can be a tool for evaluating memorization rates and quantifying potential privacy risks of representation learning models.
\vspace*{-0.3cm}
\paragraph{Contributions.} To summarize, our main contributions are as follows:
\begin{enumerate}[leftmargin=*]
    \item We develop simple and efficient methods to quantify dataset-level correlations for both image-only and vision-language representation learning models. Our methods enable \dejavu memorization tests without training two models on disjoints splits of the training set.
    \item We validate our proposed methods by comparing them to the two-model test, and analyze the strengths and weaknesses of both tests.
    \item We evaluate the one-model \dejavu test on open-source image-only and vision-language representation models. Our test reveals that open-source models do memorize specific training samples, but overall to a lesser degree than the same model trained on smaller subsets of data.
\end{enumerate}

\section{Related Work}

A body of literature has been built around how to detect and measure memorization of large foundation models. 

The first line of work is on {\em{extraction attacks}}~\citep{carlini2019secret, carlini2021extracting}, where the goal is to extract snippets of training data from a model. These attacks typically tend to work well when models are trained on data that is duplicated many times~\citep{kandpal2022deduplicating}, and are successful on a very small fraction of training data. Consequently, it is challenging to use them to develop a consistent metric that can be used to compare different models in terms of their memorization capacity.  

The second line is on {\em{membership inference}}~\citep{shokri2017membership}, which involve a statistic, such as, a loss function or score, where low values suggest membership in the training set. State-of-the-art membership inference attacks~\cite{carlini2022membership, watson2021importance} also involve training multiple ``shadow models'' that are used to calibrate the values of the statistics. Membership inference tests have close connections to overfitting~\cite{Yeom18} in that the statistic is chosen to be one that is overfitted during training. The challenges of membership inference tests is that low values of the statistic only {\em{suggest}} membership, and do not necessarily provide concrete sample-level evidence. Additionally, they sometimes do not work well on large models~\citep{duan2024membership}. Finally, we note that for representation learning methods such as DINO~\citep{Dino} that use self-distillation, loss minimization is not usually the training objective -- which might lead to failure of membership inference attacks based on loss statistics~\citep{liu2021encodermi, he2021quantifying}. In contrast, our measurement method is more concrete, agnostic to the method of training, and does not require training multiple similar models. 

A third line of work is on {\em{attribute inference}}~\citep{fredrikson2014privacy}, where we are given a model, and some attributes of a training data point, and the goal is to use the model to infer the rest.\cite{jayaraman2022attribute} recently show that most attribute inference tests apply equally well to training and test data, and hence may not be very relevant in measuring privacy. In contrast, \dejavu memorization specifically looks at sample-level attribute inference in training data points {\em{beyond what could be achieved through dataset level correlations}}, which justifies its relevance.

Our work also has connections to prior work on measuring memorization in classification models and influence functions~\citep{koh2017understanding}. \citet{feldman2020does} proposes a stability-based definition of memorization, where a classifier memorizes the label of $(x, y)$ if $f_S(x) = y$, where $f_S$ is trained on a training set $S$, and $f_{S \setminus (x, y)}(x) \neq y$ where $f_{S \setminus (x, y)}$ is trained on $S \setminus \{ (x, y)\}$. Unfortunately this can be highly computationally demanding, as measuring memorization for a single example requires training a full model. 

Stability-based memorization is also very related influence functions~\citep{koh2017understanding}, which approximate the impact of a single training example on a test prediction. Specifically, if a training point has high influence on its own prediction, then it is likely memorized. However, calculating influences, while easier than re-training a model, is also compute-heavy for large models, and involve many approximations. In contrast, our approach has lower computational cost.

\section{Measuring Dataset-level Correlations}\label{sec:dataset_level_corr}

As explained before, the main challenge with prior work is that we need a second model trained on similar data to determine if the task could be done by dataset-level correlations. Our main contribution is to introduce alternative approaches for inferring dataset-level correlations, and empirically demonstrate that these approaches suffice for the purpose of measuring memorization. 

\subsection{Formal Definition}
\label{sec:definitions}
Formally, we define memorization as follows. We have a training dataset $D = \{z_1, \ldots, z_n\}$ drawn i.i.d from an underlying data distribution $\calD$; this is used to train a representation learning model $f$. Suppose that a data point $z$ drawn from $\calD$ can be written as: $z = (v, t)$ where $v$ and $t$ represent disjoint but possibly correlated information. For example, $v$ could be the background of an image, and $t$ the label of the foreground object in it. Similarly, suppose that we can write the data distribution $\calD$ as the product of the marginal $\mu(v)$ over $v$ and the conditional distribution $\mu(t | v)$. 

Loosely speaking, \dejavu memorization happens when we can use $f$ to infer $t$ from $v$ for points $z_i$ in the training set $D$ better than what we could do from knowing $\mu(t|v)$. Formally, for a discrete label $t$, we can rigorously define memorization as follows.

\begin{definition}[\Dejavu Memorization]
\label{def:dejavu}
Let $z = (v, t)$ be a training data point. $z$ is said to be memorized if there exists a predictor $h$ such that $h(f, v) = t$ while $\argmax_{t'} \mu(t' | v) \neq t$.
\end{definition}

Observe that the second part of the definition precludes inference through dataset-level correlations. \autoref{fig:intro_figure} shows a concrete example. Suppose we have an image $z = (v, t)$ of a patch of water $v$ (background) with a black swan $t$ in the foreground. Suppose also that based on the representation $f(v)$, we can predict there is a black swan in the image. Is this image memorized by $f$? It is possible, but it might also be possible that all patches of water in the training dataset are adjacent to black swans, \emph{i.e.} $\argmax_{t'} \mu(t' | v) = t$. Therefore to determine if the image is being memorized, we need to rule out this possibility.

Related to, but different from us, \cite{feldman2020does} provides a stability-based definition of memorization. We adapt their original definition to our setting as follows.

\begin{definition}[Stability-based Memorization]
Let $z_i = (v_i, t_i)$ be a training data point, and let $f_D$ denote a model trained on the dataset $D$. $z_i$ is said to be stably memorized if there exists a predictor $h$ such that $h(f_D, v_i) = t_i$ and $h(f_{D \setminus z_i}, v_i) \neq t_i$.
\end{definition}

In other words, if we exclude $z_i$ from the training set, then we cannot use the model $f$ to predict $t_i$ correctly. Observe that this definition is very closely related to the notion of stability in learning theory~\citep{bousquet2002stability}. 

These two definitions are related, but subtly different -- there can be examples that are \dejavu memorized, but not stably memorized and vice-versa. It can be easily shown that the rate of \dejavu\ memorization is upper-bounded by the generalization error of $h$; on the other hand, the rate of stability-based memorization is by definition the leave-one-out error~\citep{bousquet2002stability} of $h$. Classical learning theory~\citep{bousquet2002stability} predicts that the leave-one-out error of a classifier is close to its generalization error. Therefore, the rates of these two notions are close for well-generalized classifiers that adapt themselves to the data such as neural networks. 

\cite{feldman2020does} provides a method to measure stability-based memorization for a sub-sample of the training data that involves training a large number of auxiliary models; in contrast, our notion has the advantage that it can be measured in a much more computationally efficient manner. 

\subsection{Image Representation Learning Models}\label{sec:vision}

For image representation learning, \dejavu memorization measures the accuracy of inferring the foreground object given a background crop. Let $\mathsf{crop}$ be a function that, when given any image $x$, produces a background crop $\mathsf{crop}(x)$. Then for a sample $z_i = (x_i, y_i) \in D$ where $x_i$ is an image and $y_i$ is the label of the foreground object, we have $v_i = \mathsf{crop}(x_i)$ and $t_i = y_i$ in the notation of Definition \ref{def:dejavu}.
Observe that since we are looking at unsupervised representation learning, the label $y_i$ was not used to train the model $f$. Define
\begin{equation}
    \mathsf{acc}_f(v, t) = \mathds{1}((h \circ f)(v) = t) \in \{0,1\},
\end{equation}
where $h$ is a predictor that takes the representation of $f(v)$ and outputs a foreground object label. Observe that $\mathsf{acc}_f(v, t)$ is a $0/1$ value which is $1$ when the foreground prediction is correct. Since $v_i$ does not contain the foreground object, for a training sample $z_i$ that is not memorized, one expects $\mathsf{acc}_f(v_i, t_i)=0$, except by sheer chance.  However, dataset-level correlations may in fact allow accurate prediction of the foreground object from a background crop, \emph{e.g.} if the foreground object is a basketball and the background is a basketball court. To isolate this effect, \citet{Dejavu} proposed to split the training set $D$ into disjoint sets $A$ and $B$, and train two models $f_A$ and $f_B$ on the two datasets. Then, for $z_i \in A$, if $\mathsf{acc}_{f_A}(v_i, t_i) = 1$ but $\mathsf{acc}_{f_B}(v_i, t_i) = 0$, one can then infer that $t_i$ cannot be predicted from $v_i$ from correlation alone, and thus $f_A$ has likely memorized $z_i$.

In this analysis, $\mathsf{acc}_{f_B}$ determines if the foreground object can be predicted from $\mathsf{crop}(x_i)$. To enable \dejavu memorization measurement with a single model, we propose to replace $\mathsf{acc}_{f_B}$ with the prediction of a reference model that directly classifies the foreground object given the background crop. We propose two ways to do this: training an image classification network end-to-end, and using naive Bayes classifier on top of an object detector.

\paragraph{Image classification network.} Our first approach is straightforward: we train an image classifier to predict $t_i$ directly given $v_i = \mathsf{crop}(x_i)$. For ImageNet, we train a ResNet50 model over a split $D'$ of the training set $D$ and evaluate \dejavu memorization on $D \setminus D'$. This ensures the reference model itself is not memorizing, but rather predicting the correlation between the background crop and the foreground object.

\paragraph{Naive Bayes classifier.} If the training set $D'$ for the image classifier is large, the above approach can be just as expensive as training the model $f_B$. Our second approach alleviates this by fitting a simpler model, a naive Bayes classifier, on top of objects detected in $\mathsf{crop}(x)$. In detail, let $\mathsf{objects}$ be an object detection model with vocabulary set $\mathcal{V}$; that is, for a given image $x$, $\mathsf{objects}(x) \in \{0,1\}^{|\mathcal{V}|}$ is a binary vector such that $\mathsf{objects}(x)_k=1$ if and only if object $o_k$ exists in image $x$ for each $o_k$ in the vocabulary set $\mathcal{V}$. We then derive the empirical probability estimates over a split $D'$ of the training set:
$$P(o_k) = \frac{1}{|D'|} \sum_{z_i \in D'} \mathsf{objects}(v_i)_k, \quad P(o_k ~|~ t_i = t) = \frac{1}{|\{z_i \in D' : t_i=t\}|} \sum_{z_i \in D' : t_i = t} \mathsf{objects}(v_i)_k.$$
For a sample $z_i \in D \setminus D'$, the naive Bayes classifier predicts the probability $P(t_i=t ~|~ v_i)$ for each foreground object $t$ given the background $v_i$ as:
\begin{align*}
    P(t_i=t ~|~ v_i) &= P(t_i=t ~|~ \text{ detected objects in } v_i) = P(t) \prod_{k : \mathsf{objects}(v_i)_k > 0} \frac{P(o_k ~|~ t_i=t)}{P(o_k)},
\end{align*}
where the last equality uses the independence assumption for naive Bayes. In practice, because the object detection result can be noisy, we truncate the list of detected objects to the top-$K$ according to detection score.

\begin{figure*}[t]
    \centering
    \begin{subfigure}{.5\textwidth}
        \captionsetup{width=0.9\textwidth}
        \includegraphics[width=\textwidth, height=6cm]{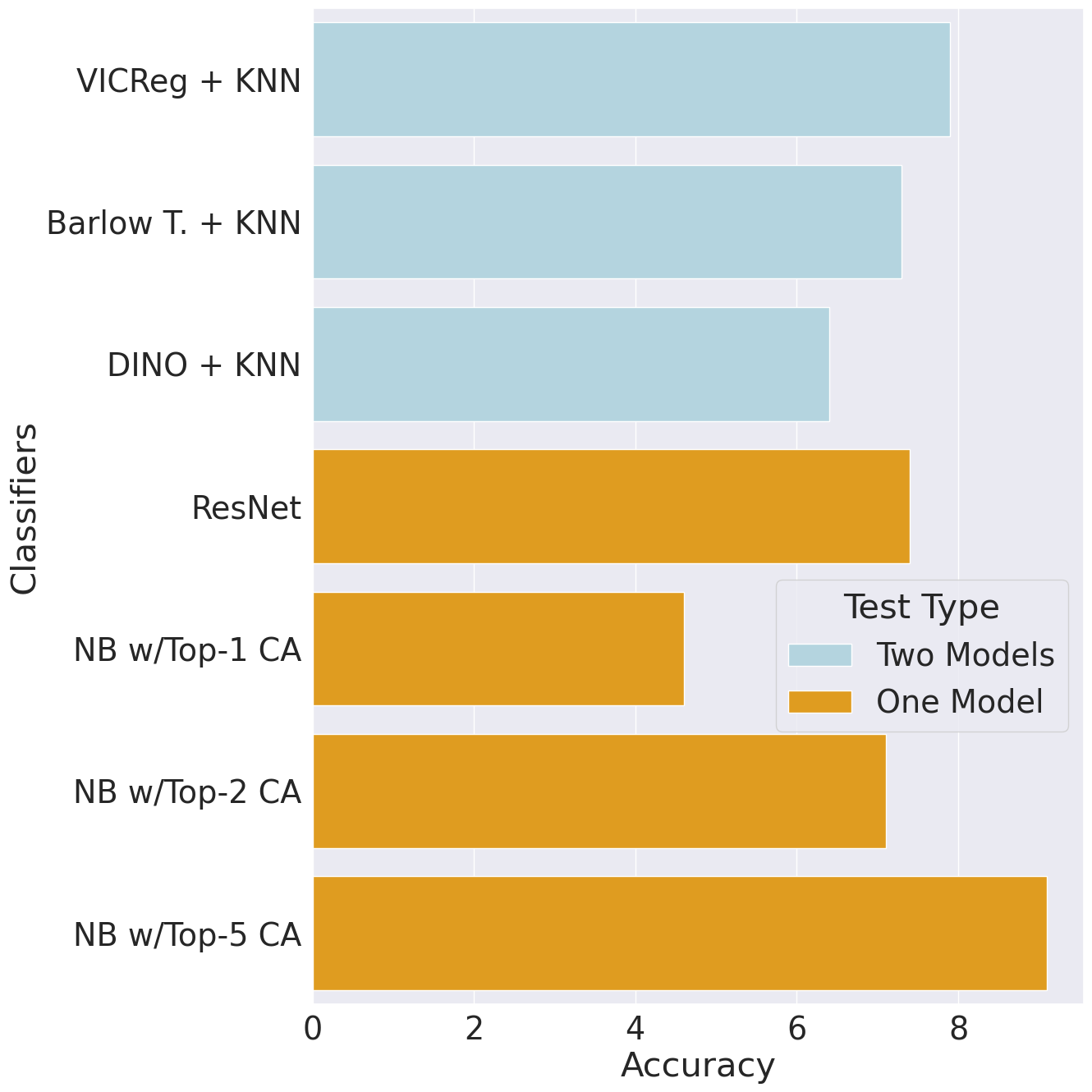}
        \caption{
        Aggregate accuracy(\%) of foreground prediction from background for different models. We see that ResNet50 and NB Top-2 are similar to both VICReg and Barlow Twins, and the aggregate accuracy is low.}
        \label{fig:corr_pop_acc}
    \end{subfigure}%
    \hspace{2ex}
    \begin{subfigure}{.47\textwidth}
        \captionsetup{width=0.9\textwidth}
        \includegraphics[width=\textwidth, height=6cm]{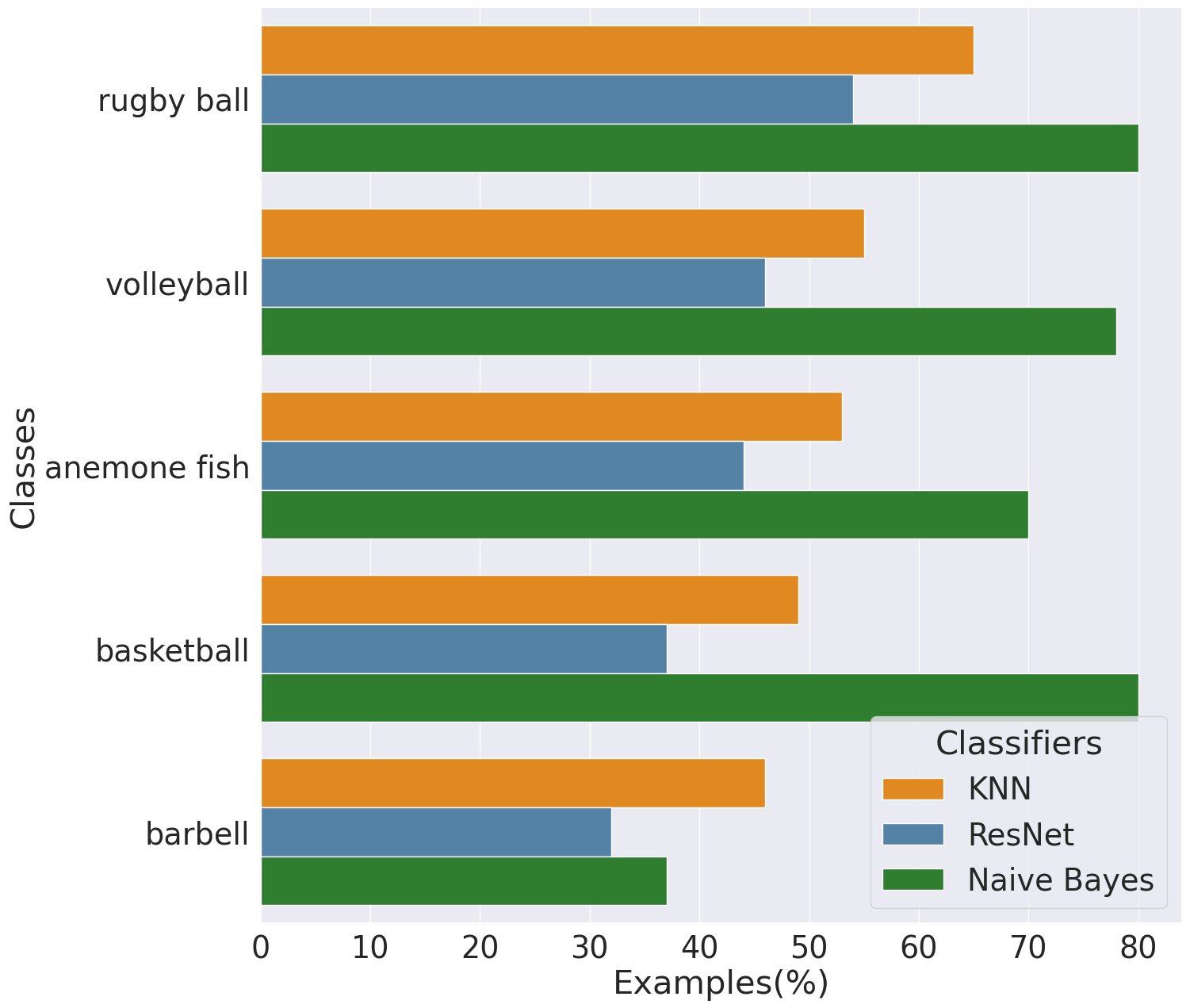}
        \caption{\protect Accuracy of the top-5 predicted classes based on dataset-level correlations using three classifiers: KNN, Resnet and Naive Bayes. Naive Bayes classifier uses top-20 crop annotations as features. }
        \label{fig:top5_correlations}
    \end{subfigure}
    \caption{Left: Population-level correlation accuracy scores across different models. The accuracies for two model tests are based on KNNs computed on top of VICReg, Barlow Twins and DINO representations. ResNet50 and Naive Bayes classifier are used for one model tests. The results show that ResNet50 and NB Top-2 are similar to both VICReg and Barlow Twins. Right: Corresponding Top-5 predicted dataset-level correlation classes and the percentage of per class correlated examples. }
    \label{fig:vision_heatmap_classes}
    \vspace{-2ex}
\end{figure*}

\begin{figure*}[t]
    \centering
    \begin{subfigure}{.47\textwidth}
        \captionsetup{width=0.9\textwidth}
        \includegraphics[width=\textwidth, height=5cm]{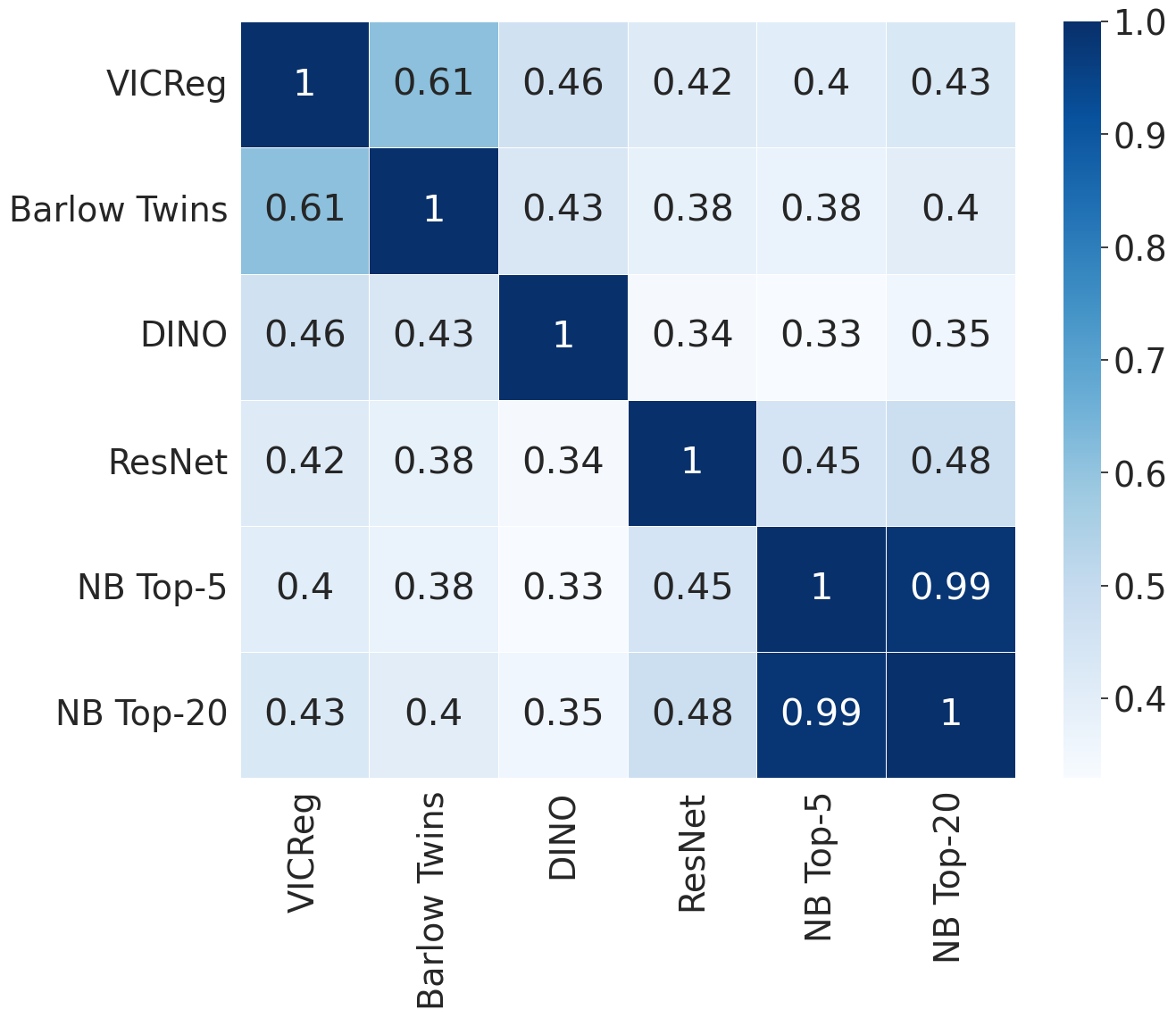}
        \caption{Pairwise sample-level correlation agreement fraction among six reference models. VICReg, Barlow Twins and DINO are used for two model tests whereas ResNet50, NB Top-5 and NB Top-20 for one model tests.}
        \label{fig:heatmap_corr}
    \end{subfigure}%
    \hspace{2ex}
    \begin{subfigure}{.47\textwidth}
        \captionsetup{width=0.9\textwidth}
        \includegraphics[width=\textwidth, height=5.4cm]{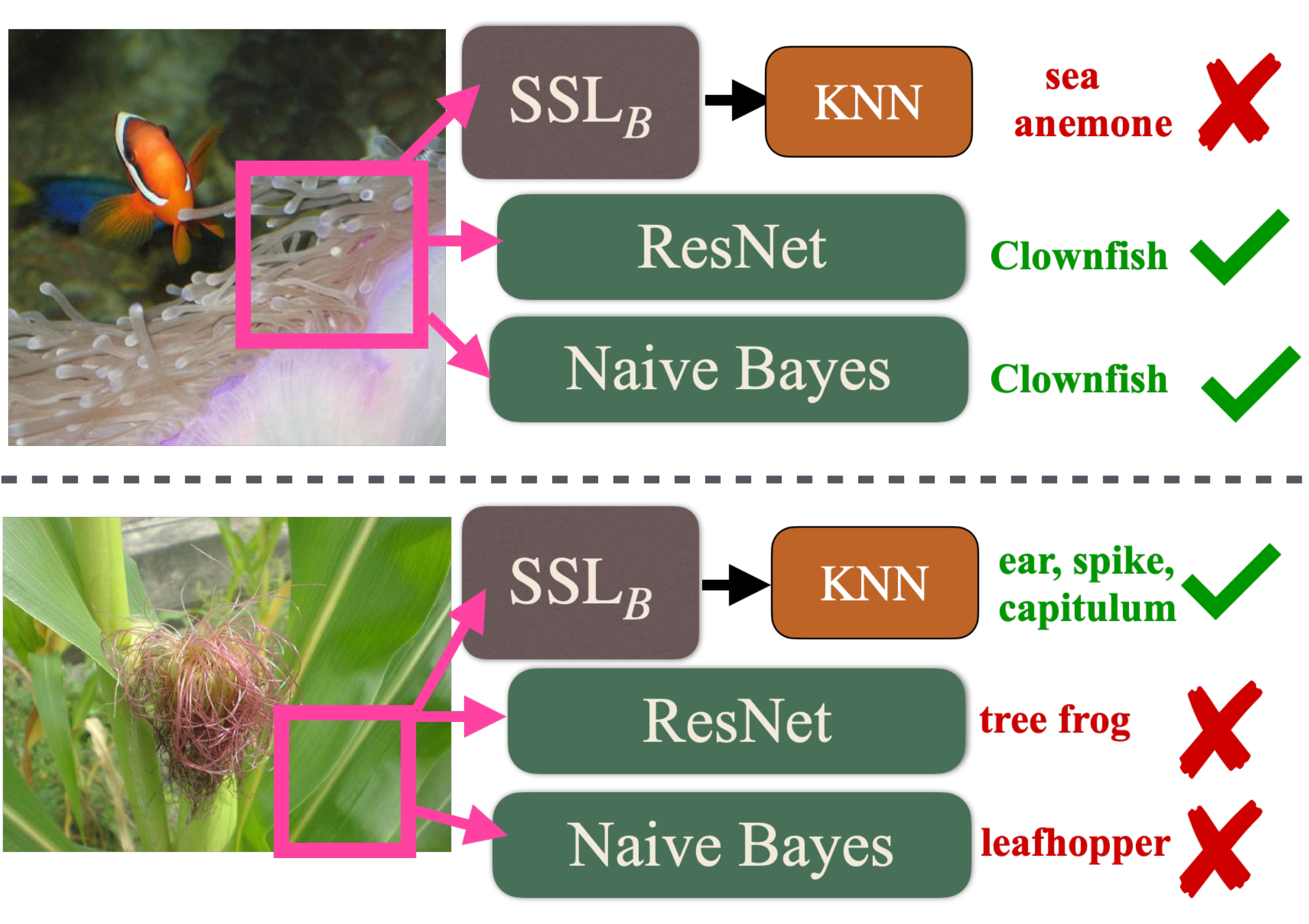}
        \caption{Examples demonstrating when one model tests (ResNet and Naive Bayes classifiers) succeed and two model tests (KNN) fail and vice versa.}
        \label{fig:examples_corr}
    \end{subfigure}
    \caption{Left: Pairwise sample-level agreement in measuring dataset-level correlations and Right: Examples demonstrating when one model tests (Resnet and Naive Bayes classifiers) succeed and two model tests (KNN) fail and vice versa. One model tests learn the correlations between foreground and background better since it is enforced by the classifier training, however, they are less accurate when the relationships between foreground and background are ambiguous. One model tests, in contrast, are better at disambiguating the foreground and background relationships. They, however, sometimes tend to predict what's on the background and not what foreground it is associated with. }
    \label{fig:vision_heatmap_classes}
    \vspace*{-0.2cm}
\end{figure*}

\paragraph{Results.} We now investigate how effective the two approaches are at measuring dataset-level correlations in comparison with a second model~\citep{Dejavu}.
Specifically, we compare the ResNet classifier and two versions of the Naive Bayes Classifier that uses the top-5 and top-20 crop annotations, as well as three SSL models---VICReg~\cite{vicreg}, DINO~\cite{Dino} and Barlow Twins~\cite{zbontar2021barlow} ---and look at how much these classifiers agree on the predicted correlations.

\autoref{fig:corr_pop_acc} shows that the overall accuracy across these classifiers are largely comparable. We then zoom into top-5 most correlated classes in \autoref{fig:top5_correlations}, where we show the number of correctly predicted correlations for the top-5 most correlated classes. Across the three methods that we compared, namely KNN, ResNet and Naive Bayes, the top-5 most correlated classes are identical.
However, at a sample level, there is in fact a large divergence in prediction across different methods. \autoref{fig:heatmap_corr} shows the fraction of samples where the correlation prediction agreed for the different reference models. Here, we see that the agreement is quite low, only about $40\%$.
This suggests that the methods have different inductive biases from the SSL-based classifiers when measuring dataset-level correlations and thus can overestimate memorization when used in the one-model test.
Appendix~\ref{appendix:singlevstwomodel} looks deeper into the intersection of common memorized examples across multiple reference models. It shows that ResNet classifier agrees with the intersection of three two model tests for approximately 86\% and Naive Bayes for 78\% of top-1 correlated examples.
Figure~\ref{fig:examples_corr} showcases different scenarios when the reference models agree and disagree. It unveils the strengths and the weaknesses of the one and two model tests and suggests that these methods can be used conjointly.

\subsection{Vision Language Models}\label{sec:vlm}


For vision-language models (VLMs), the training dataset $D$ consists of image-text pairs $z = (z_\text{img}, z_\text{text})$. The model $f$ learns to simultaneously embed $z_\text{img}$ and $z_\text{text}$ into low-dimensional representations, with the training objective of aligning the representations $f(z_\text{img})$ and $f(z_\text{text})$. Following the setup of \cite{jayaraman2024deja}, we consider $v=z_\text{text}$ and $t=\mathsf{objects}(z_\text{img}) \in \{0,1\}^{|\mathcal{V}|}$, where $\mathsf{objects}(z_\text{img})$ is the set of detected objects in a vocabulary $\mathcal{V}$. \Dejavu memorization occurs when one can leverage $f$ to infer objects in $z_\text{img}$ using $z_\text{text}$ \emph{significantly beyond dataset-level correlation}. Specifically, consider a predictor $h$ that operates on $f(v)$ and outputs a binary vector of predicted objects. We can define the precision and recall metrics for the predictor:
\begin{equation}
    \label{eq:prec_and_recall}
    \mathsf{prec}_f(v, t) = \frac{\langle (h \circ f)(v), t \rangle}{\| (h \circ f)(v) \|} \in [0,1], \quad \mathsf{recall}_f(v, t) = \frac{\langle (h \circ f)(v), t \rangle}{\| t \|} \in [0,1].
\end{equation}
One might expect $\mathsf{prec}_f(v, t) = \mathsf{recall}_f(v, t) = 0$ when $f$ does not memorize. However, dataset-level correlations may in fact enable the prediction of objects in $z_\text{img}$ from $z_\text{text}$, \emph{e.g.} if $z_\text{text} = \texttt{A table full of fruits and vegetables}$ and $z_\text{img}$ contains objects such as apples, oranges, carrots, \emph{etc.} To design one-model \dejavu memorization tests, we would like to capture this type of dataset-level correlation with a reference model.
This is especially hard for VLMs since these models are typically trained on internet-scale datasets consisting of billions of diverse samples under a long-tailed distribution. Training this reference model from scratch on a subset of $D$ requires a similar effort as training the VLM itself, which defeats the purpose of a one-model test.

\paragraph{Using pre-trained text embedding models as reference models.} To tackle this challenge, we leverage a pre-trained text embedding model $g$ that transforms text into vector representations, with the requirement that $\langle g(z_\text{text}), g(z_\text{text}') \rangle$ is high when $z_\text{text}$ and $z_\text{text}'$ are semantically similar (and vice versa). We can then utilize $g$ to define a reference model similar to the two-model setup of \citet{jayaraman2024deja}. Given a public set $D_\text{pub}$ of image-text pairs and a training sample $z$, the reference model first performs inner product search in the embedding space of $g$ to find the $K$ most similar captions in $D_\text{pub}$, $(z_1')_\text{text},\ldots,(z_K')_\text{text}$.
Then, we predict $o_k \in z_\text{img}$ if and only if $o_k \in (z_j')_\text{img}$ for some $j \in \{1,\ldots,K\}$; see \autoref{fig:ss_example} in Appendix \ref{sec:vlm_additional_res} for an example.

\paragraph{Result.} We investigate how well the LLM ($g$) captures the dataset-level correlations for predicting ground-truth objects in images when compared to the reference VLM ($f_B$) of \cite{jayaraman2024deja}, that has not seen the target images in its training. 
\begin{wrapfigure}{r}{0.6\textwidth}
    \centering
    \vspace{-2ex}
    \begin{subfigure}{0.28\textwidth}
    \centering
        \includegraphics[width=\linewidth]{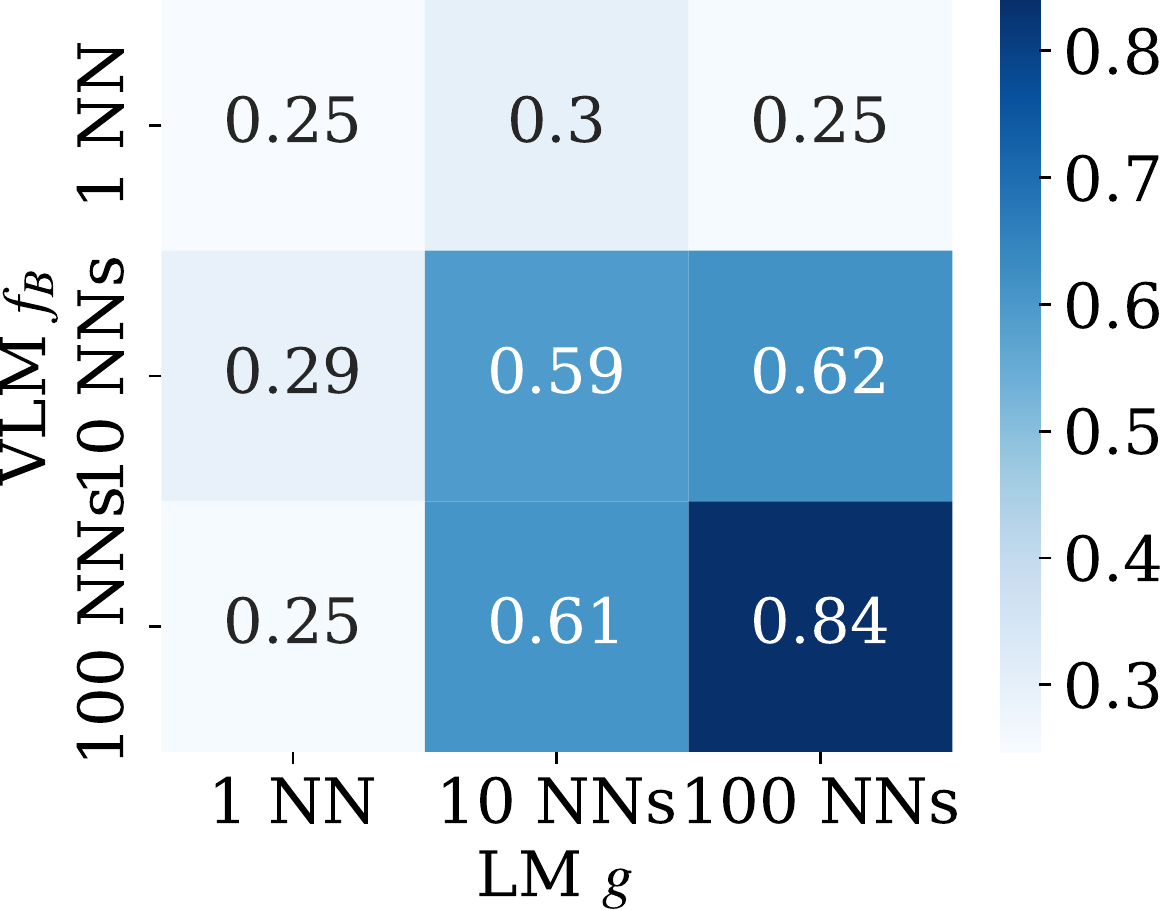}    
        \label{vlm_jaccard_nns}
        \vspace{-2ex}
        \caption{Predicting all objects with varying NNs}
    \end{subfigure}\hfill
    \begin{subfigure}{0.28\textwidth}
    \centering
        \includegraphics[width=\linewidth]{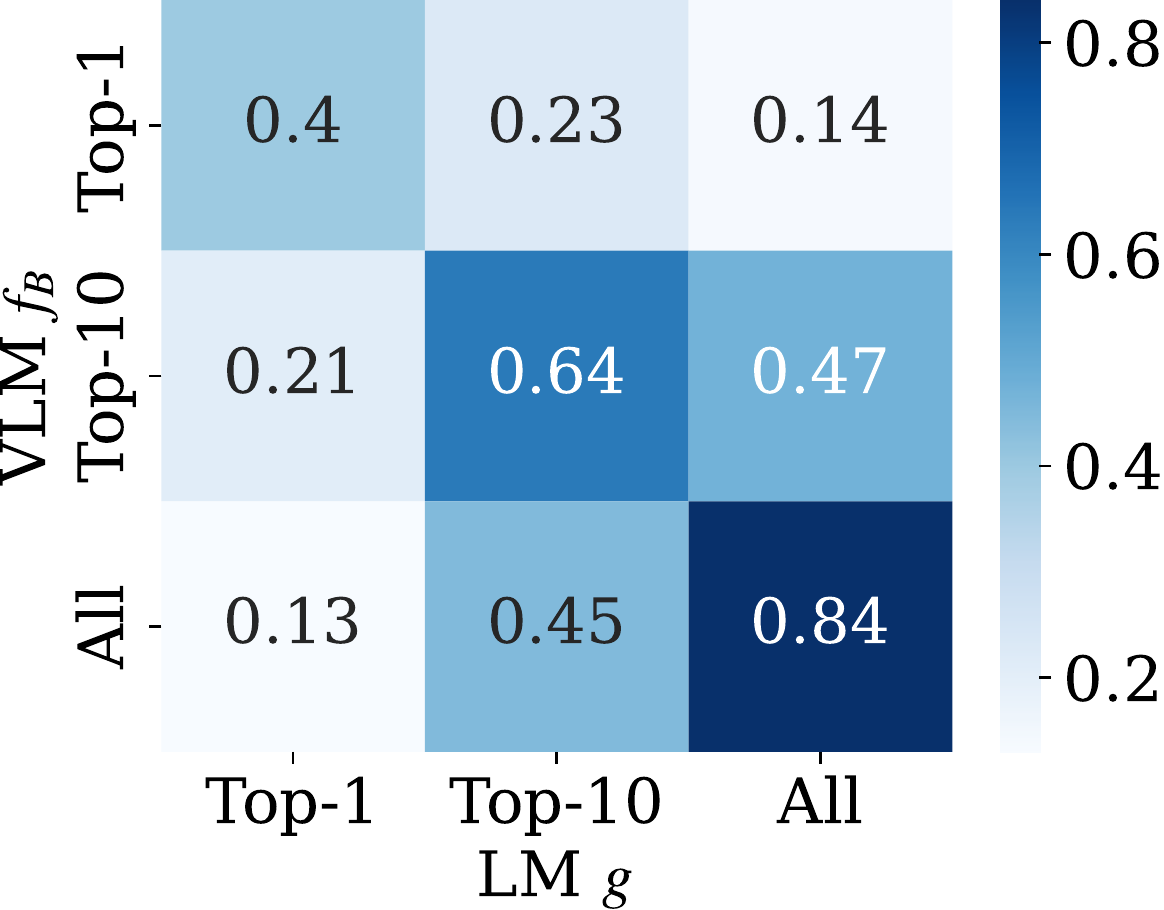}   
        \label{vlm_jaccard_labels}
        \vspace{-2ex}
        \caption{Predicting top-$k$ objects with 100 NNs}
    \end{subfigure}
    \caption{Pairwise sample-level agreement (using Jaccard similarity for predicting correct objects) between the reference VLM $f_B$ in previous two-model test and the GTE language model $g$. The heatmap shows that the agreement fraction for one model and two model tests are comparable. 
    }
    \label{fig:vlm_jaccard}
    \vspace{-2ex}
\end{wrapfigure}
We plot heatmaps for pairwise sample-level agreement similar to the vision model case above. However, since this setting has multiple objects per image, we calculate the Jaccard similarity between the correct object predictions per sample for the two models $g$ and $f_B$, and report the averaged value across all the training samples. 

\autoref{fig:vlm_jaccard} shows the pairwise sample-level agreement between the two models for predicting various top-$k$ object labels and for different number of NNs. As shown, even when predicting all objects, the two models agree only on 84\% objects on average. This agreement decreases as we limit the number of top-$k$ object predictions or alternatively limit the number of NNs. We see a similar trend that reference models do not always agree on the predictions.  
We show some examples of what the two models, VLM $f_B$ and LLM $g$, predict for a given caption in \autoref{fig:vlm_corr_examples} in the appendix.

\section{Measuring \Dejavu Memorization using One Model Test}\label{sec:vision_vlm_mem}

In this section, we investigate how effective new methods for measuring dataset-level correlations are when we use them for measuring memorization. Specifically, we look at two main questions:
\textbf{1.} How close are the results of the single-model deja-vu test to the two-model test?
\textbf{2.} What is the fraction of memorization in open-source (OSS) pre-trained representation learning models?
These questions are addressed in the context of both image representation learning models and vision language models. 

\subsection{Image Representation Learning}\label{section: vision_exp}


\textbf{Dataset.} We conduct all our image representation learning experiments on ImageNet~\cite{imagenet} dataset.

We use 300k (300 per class) examples to train the reference models to learn dataset-level correlations. We measure memorization accuracy on an additional disjoint set of 300k images. For the two model tests, these images are included in the training set of the target models, but not the reference models. Finally, we use another additional distinct 500k images to predict the nearest foreground object given the representation of a background crop through KNN.

\textbf{Models.}
Two model tests are conducted analogous to~\cite{Dejavu}.
One model tests rely on a classifier that is trained once to predict dataset-level correlations for SSL models. The dataset used to train this classifier overlaps with the training dataset of open-source models but is disjoint from the subset of the examples for which we measure memorization. 


We compare two kinds of classifiers to detect dataset-level correlations. The first is a ResNet50 trained on the background crops to predict the foreground object. We used LARS optimizer and 0.1 weight decay for L2 regularization to avoid overfitting.
The second classifier is a Naive Bayes classifier. It uses background crop annotations as features. We automatically annotate background crops using Grounded-SAM~\citep{liu2023grounding, ren2024grounded}. Annotations represent textual tags associated with probability scores. We use these probability scores to pick top 1, 2 and 5 features to compute final Naive Bayes probability scores.
The reference models are trained on a single machine with 8 Nvidia v100 GPUs, 32GB per GPU using 128 batch size. All other experiments are performed on the same machine. 

\textbf{Metrics.} Following~\cite{Dejavu}, we report the \dejavu score and the \dejavu score at $p\%$. The \dejavu\ score for a model $f$ is the difference between two accuracy values: the first is the accuracy of predicting the foreground label $y$ from the representation $f(v)$ of the background crop $v$ based on KNN. The second is the accuracy of predicting $y$ from a reference model. The \dejavu score at $p\%$ is the difference between the same two accuracies, but now calculated only on the top $p\%$ of the most confident examples.

\subsubsection{How close is the \dejavu memorization of one-model and the two-model tests?}

\begin{figure}[h]
    \centering
    \vspace{-2ex}
    \begin{minipage}[c]{0.55\linewidth}
        \includegraphics[width=\linewidth, height=0.6\textwidth]{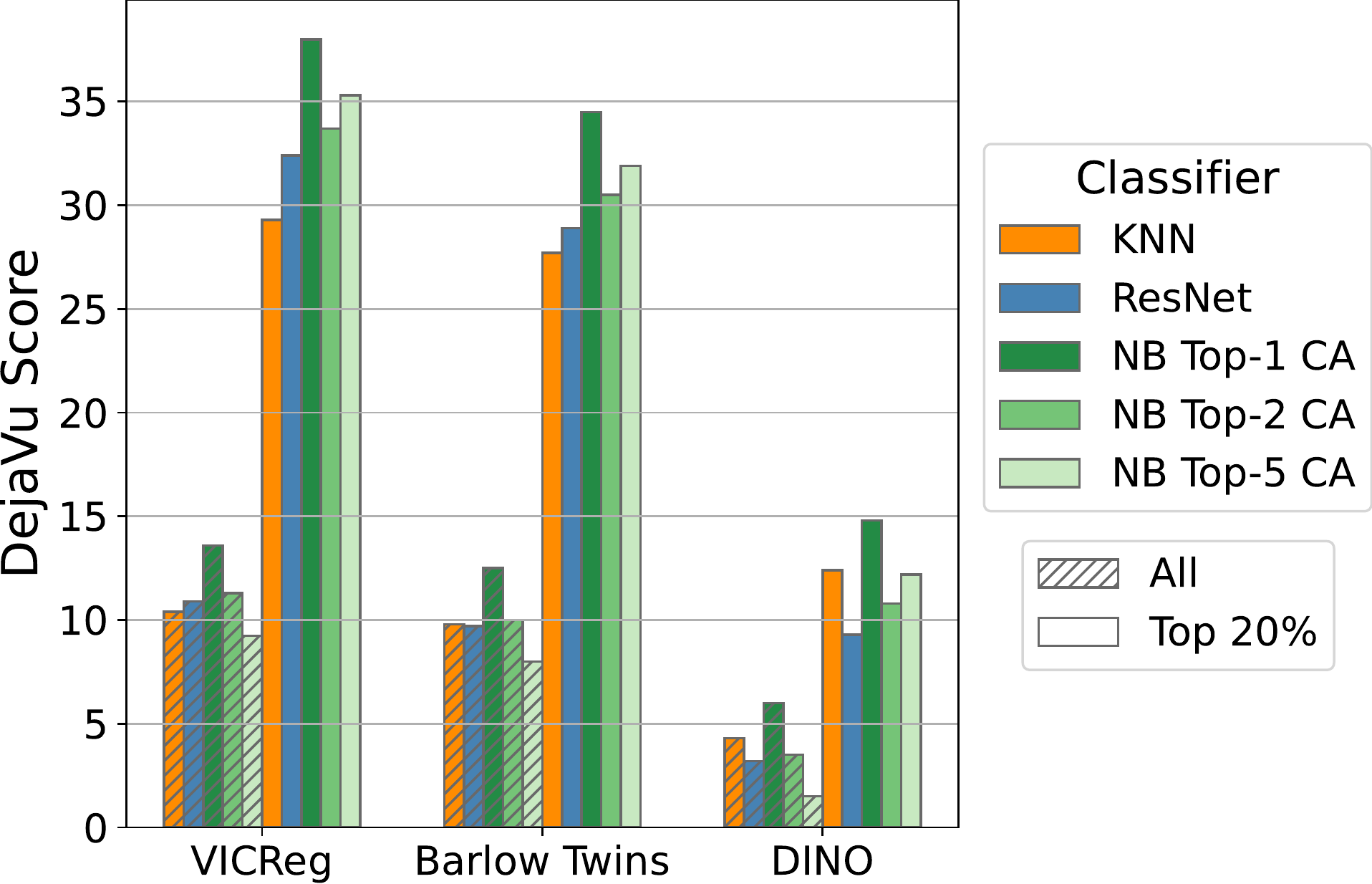}
        \caption{Comparison of overall and Top 20\% most confident \Dejavu (DV) scores using one model (ResNet Classifier, Naive Bayes w/ Top-k Crop Annotations (CA)) and two model (KNN Classifier) tests for VICReg, Barlow Twins and DINO trained on a 300k subset of ImageNet.}
        \label{fig:dejavu_ref_dv_vicreg_barlow_dino}
    \end{minipage}
    \hfill
    \begin{minipage}[c]{0.40\linewidth}
        \includegraphics[width=\linewidth, height=0.85\textwidth]{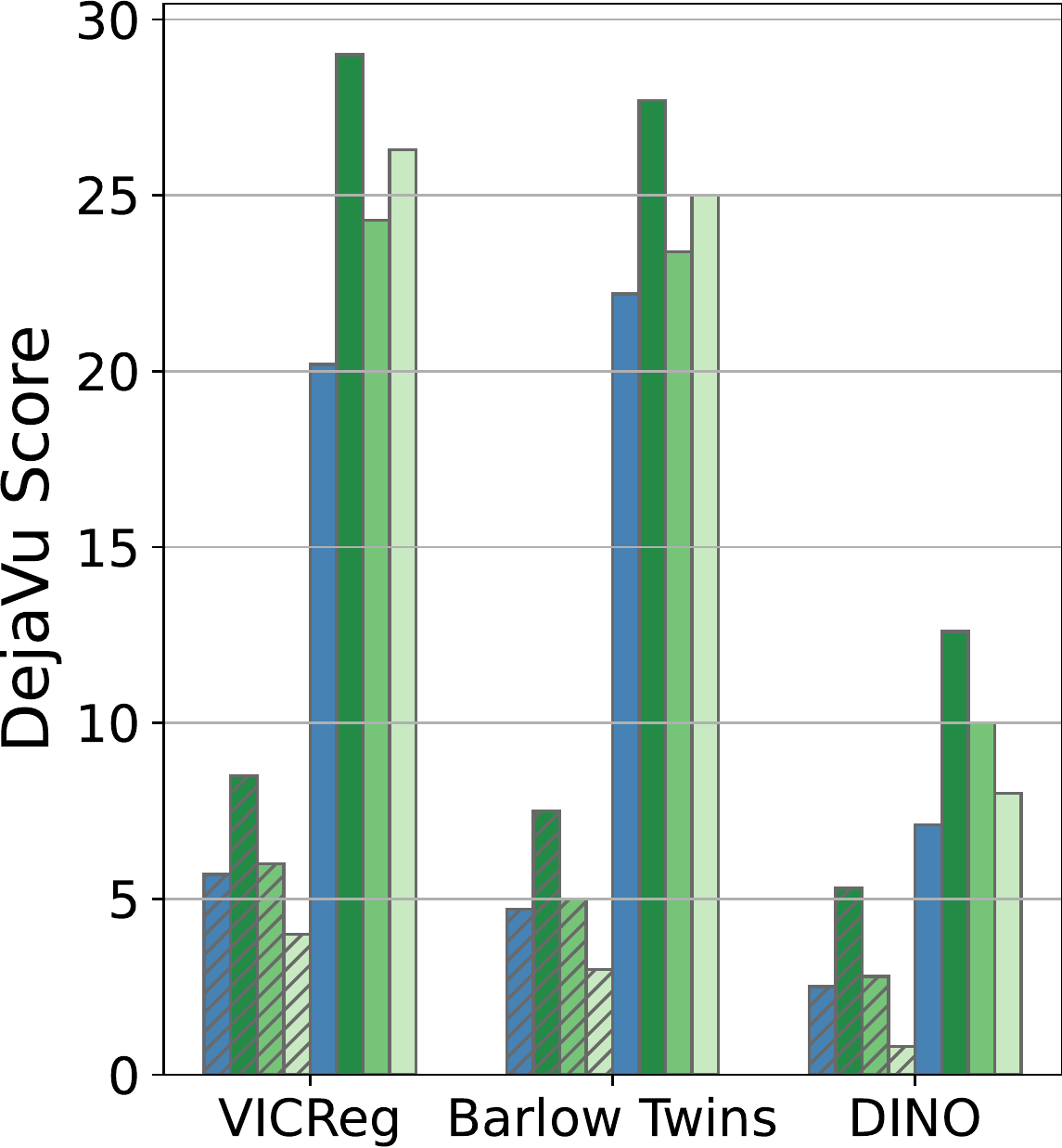}
        \caption{Comparison of overall and Top 20\% most confident \Dejavu (DV) scores using one model (ResNet Classifier, Naive Bayes w/ Top-k Crop Annotations (CA)) tests for pre-trained VICReg, Barlow Twins and DINO.}
        \label{fig:dejavu_dv_oob_vicreg_barlow_dino}
    \end{minipage}%
    \vspace{-2.2ex}
\end{figure}

Section~\ref{sec:vision} discusses how close one and two model tests are in terms of dataset-level correlation accuracy. In this section we compare \dejavu memorization scores for one and two model tests. \autoref{fig:dejavu_ref_dv_vicreg_barlow_dino} shows that KNN classifier (two model test) and ResNet classifier (one model test) identify similar amount of \dejavu memorization for VICReg~\citep{vicreg} and Barlow Twins models. \Dejavu memorization is substantially lower in DINO~\citep{Dino}. Similar findings are reported in ~\cite{Dejavu} as well. In addition, we observe that \dejavu score decreases as we increase the number of features (crop annotations) in Naive Bayes.
This is due to the increasing accuracy of dataset-level correlation as we increase the number of crop annotations.

\subsubsection{Do pre-trained representation learning models in the wild exhibit \dejavu memorization?}
In this section we present \dejavu memorization for pre-trained OSS representation learning models on population-level using one model tests. Two model tests aren't applicable in this scenario since pre-trained models are trained on the entire ImageNet dataset and the validation dataset is relatively small to be considered for training a second representation learning model.

Hence, \autoref{fig:dejavu_dv_oob_vicreg_barlow_dino} compares only one model tests. A comparison of one model tests between \autoref{fig:dejavu_ref_dv_vicreg_barlow_dino} and \autoref{fig:dejavu_dv_oob_vicreg_barlow_dino} shows that pre-trained models memorize less compared to the same models trained on a smaller subset of the training data. We hypothesis that this is due to the lower generalization error of the pre-trained models as a result of having a larger training set.
We provide additional examples of common dataset-level correlations and memorized images in appendix \autoref{sec:common_mem_corr}

\subsubsection{Sample-level memorization}
\begin{wrapfigure}{p}{0.76\textwidth} 
    \centering
    \vspace{-3ex}
    \includegraphics[height=0.5\textwidth] 
    {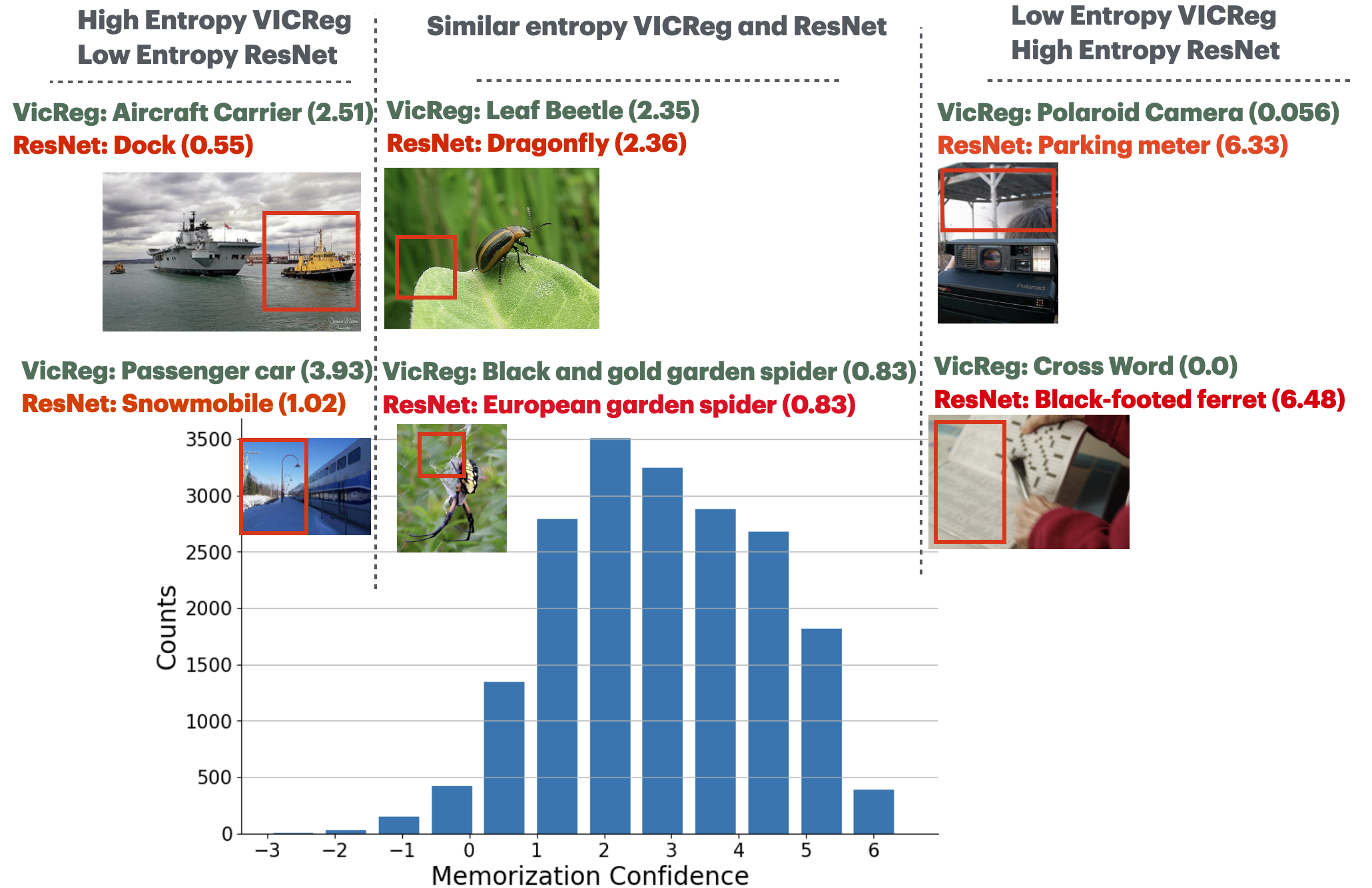}
    \caption{A histogram of sample-based memorization confidence for VICReg OOB model. Given a background patch, VICReg predicts the correct class (green). ResNet (correlation classifier) predicts the incorrect (red) class. 
    } 
    \label{fig:mem_confidence_vision}
    \vspace{-5ex}
\end{wrapfigure}

Figure~\ref{fig:mem_confidence_vision} visualizes the distribution of memorization confidence scores for pre-trained VICReg OSS model with ResNet as correlation detector. 
The memorization confidence for the $i$-th example is computed based on the following formula:
\begin{equation}
MemConf(x_i) = Entropy (Correlation\ Classifier) - Entropy_{SSL}(KNN)
\end{equation}
$Entropy_{SSL}(KNN)$ is computed according to~\citep{Dejavu}'s Section 4 description and $Entropy (Correlation\ Classifier)$ is correlation classifier's entropy over the softmax values.

~\ref{fig:mem_confidence_vision} shows that the memorized examples with high memorization confidence scores are rarer and more likely to be memorized.
The examples in the middle of the distribution are easy to be confused with another class.
E.g. Black and gold garden spider with European garden spider. On the other hand the examples with negative memorization confidence have higher memorization and slightly lower correlation entropy. 

\subsection{Vision Language Models}\label{sec:vlms}

\paragraph{Experiment setup.} We train CLIP models using the OpenCLIP~\cite{openclip} framework on the Shutterstock dataset (a private licensed dataset consisting of 239M image-caption pairs). See \autoref{sec:vlm_experiment_setup} for details on dataset preparation and training.
We quantify dataset-level memorization using the population precision gap (PPG) and population recall gap (PRG) metrics of \cite{jayaraman2024deja}. These metrics capture the population-level gap between the fraction of memorized objects and fraction of objects inferred through correlation; see \autoref{sec:vlm_experiment_setup} for details.

\begin{figure}
    \centering
    \begin{subfigure}{0.55\textwidth}
    \centering
    \includegraphics[width=\textwidth]{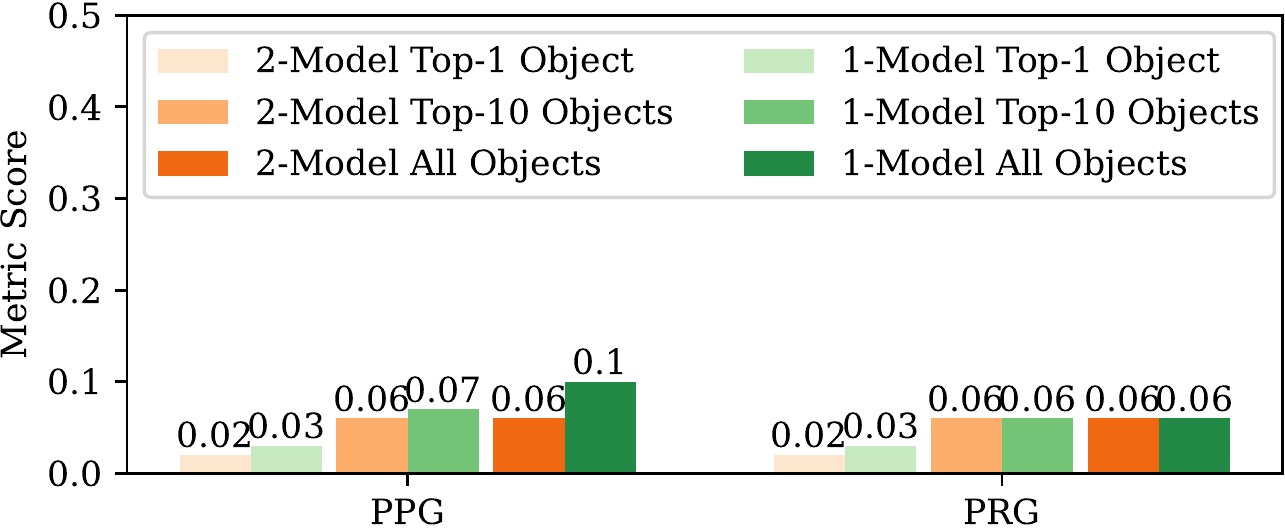}
    \caption{One-model vs two-model tests for Shutterstock models.}
    \label{fig:vlm_mem_cmp}
    \end{subfigure} \hfill
    \begin{subfigure}{0.42\textwidth}
    \centering
    \includegraphics[width=\textwidth]{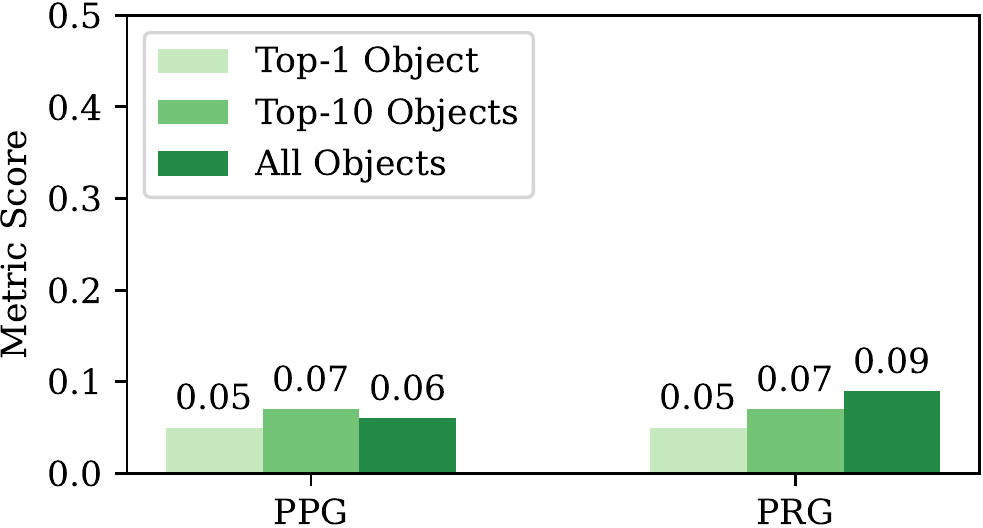}
    \caption{OSS model pre-trained on YFCC15M.}
    \label{fig:vlm_mem_oob}
    \end{subfigure}
    \caption{Data set level memorization of various VLMs. We use top-10 public set NNs to predict the top-$k$ objects and report PPG and PRG as done in \cite{jayaraman2024deja}.}
    \label{fig:vlm_mem}
    \vspace{-2ex}
\end{figure}

\subsubsection{How close is the \dejavu memorization of one-model and the two-model tests?}

As explained in \autoref{sec:vlm}, in our one-model test we use a GTE language model $g$ as a reference model to quantify data set level memorization of a target VLM $f$. We compare this test to the previous work of \cite{jayaraman2024deja}, which trains a reference VLM from scratch on a separate hold-out set. \autoref{fig:vlm_mem_cmp} compares the two tests in terms of the PPG and PRG metrics for predicting top-$k$ object labels in training images with 10 nearest neighbors from the Shutterstock public set. While the previous two-model test achieves 0.06 PPG and PRG values for predicting top-10 objects, our approach obtains 0.07 PPG and 0.06 PRG values for the same setting. Our test thus slightly overestimates the memorization as in the vision case above. We also compare the dataset-level metrics for the two tests for different settings where we vary both the number of nearest neighbors used in the test and also the number of top-$k$ objects predicted in \autoref{tab:vlm_dejavu_40M_nns} and \autoref{tab:vlm_dejavu_40M_labels} respectively in \autoref{sec:vlm_additional_res}.

\subsubsection{Do pre-trained vision-language models in the wild exhibit \dejavu memorization?}

We perform our one-model test against an out-of-the-box ResNet-50 CLIP model pre-trained on the YFCC15M data set from OpenCLIP. \autoref{fig:vlm_mem_oob} shows the PPG and PRG values for predicting different top-$k$ objects. These results are comparable to our one-model test results in \autoref{fig:vlm_mem_cmp} where we evaluate our CLIP model trained on 40M Shutterstock data. More specifically, for predicting top-10 objects with 10 nearest neighbors from public set, our Shutterstock model achieves 0.07 PPG and 0.06 PRG, whereas the OSS YFCC15M pre-trained model achieves 0.07 PPG and PRG values. 
Additional results can be found in \autoref{tab:vlm_oob_yfcc15m_nns} and \autoref{tab:vlm_oob_yfcc15m_labels} in the appendix.
We include the most memorized examples for our Shutterstock models in \autoref{fig:vlm_memorized_examples} in \autoref{sec:vlm_additional_res}.

\subsubsection{Sample-level memorization}
\begin{figure}[ht!]
    \centering
    \vspace{-2.2ex}
    \includegraphics[width=\linewidth]{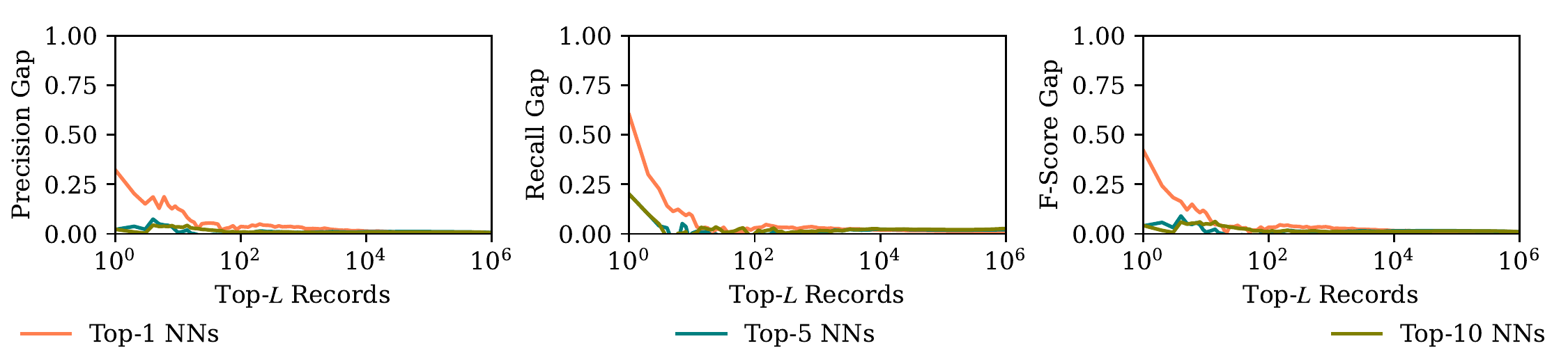}
    \caption{Sample-level memorization in VLM trained on 40M Shutterstock images, quantified in terms of precision and recall gap between target VLM and off-the-shelf GTE LM. 
    }
    \label{fig:vlm_top_all}
    \vspace{-3.5ex}
\end{figure}

Figure~\ref{fig:vlm_top_all} shows samples with higher degree of memorization. The samples are sorted from high to low memorization such that the top-L samples have higher precision and recall gaps for recovering objects using target and reference models. We find the gap between the objects recovered from target and reference models for each training record, and estimate the precision and recall gaps. A positive gap indicates that the target model memorizes the training sample and the magnitude of the gap indicates the degree of memorization. 





\section{Discussions and Conclusion}

This paper proposes a principled method for measuring memorization in vision and vision-language encoder models that does not rely on training similar shadow models. Our method enables, for the first time, direct measurement of memorization in open-source representation learning and vision-language models. It also allows us to quantify potential privacy risks associated with these models. One consequence of these new measurements is that now we can find out how much different OSS models memorize. In particular, we find that VicReg and Barlow Twins memorize more than DINO. Additionally, all standard OSS models memorize less than their versions trained on subsets of the data. 

Finally, our method of measurement involves approximations to theoretical quantities, and as such, has some limitations when these approximations do not hold. One such limitation is that our alternative dataset-level correlation estimation might be a poor approximation to the Bayes optimal, or might itself memorize its own training set, thus skewing the results. However, given that these are much simpler classifiers, their own rate of memorization is expected to be lower. Another limitation is that the additional annotations that we use for our measurements may be lower quality, which might also lead to biased results. A closer analysis of the impact of these factors is an avenue for future work.

\newpage

\section*{Acknowledgements}

We thank Maxime Oquab for pointing us to the challenges of the two-model test. 

\bibliographystyle{plainnat}
\bibliography{citations}

\newpage
\appendix
\section{License of the assets}
\label{sec:licenses}

\subsection{License for the code}
We use the code from \citet{Dejavu} which is under the Attribution-NonCommercial 4.0 International license according to \url{https://github.com/facebookresearch/DejaVu?tab=License-1-ov-file#readme}. We also use the code from \citet{ren2024grounded} which is under the Apache 2.0 licence according to \url{https://github.com/IDEA-Research/Grounded-Segment-Anything?tab=Apache-2.0-1-ov-file}.

\subsection{License for the datasets}
We use ImageNet\citep{yang2021imagenetfaces} which license can be found at \url{https://www.image-net.org/download.php}. We also use a private licensed dataset consisting of 239M image-caption pairs.

\section{Additional results for Image Representation Learning}

\subsubsection{How close are the results of the single-model deja-vu test and the two-model test?}\label{appendix:singlevstwomodel}

We observe that the intersection of correctly predicted examples between the reference models is relatively small. This intersection is approximately 40\% for two model tests. This tells us that there is significant noise in predicting dataset level correlations even if the models are trained on the same dataset. In order to better understand this phenomenon, we intersect common subsets of two model tests with one model test. Table~\ref{tab:dejavu_ref_example_test} shows that Resnet50 and Naive Bayes with Top 20 annotation tags are able to predict the same correlations for almost 60\% of the test examples that were also predicted as correlated by by two model tests. This percentage increases if we look into top-20, top-5 and top-1 predictions. For top-1 predictions Resent50 reaches over 86\%.
In addition to that we also observe that example-level correlation accuracy increases for Naive Bayes by increasing the number of features. This tells us that Naive Bayes becomes more accurate if we increase the number of features describing the crop.

\begin{table}[htbp!]
\centering
\begin{tabular}{|l|l|l|l|l|}
\hline
\thead{Intersection between \\ VICReg, Barlow Twins, DINO \\ AND}
 & \thead{Accuracy } & \thead{Accuracy \\ Top20} & \thead{Accuracy \\ Top-5} & \thead{Accuracy \\ Top-1} 
\\
\hline
NB w/Top-1 Crop Annotation & 32.04\%  & 31.12\% & 29.78\% & 13.51\%  \\
NB w/Top-2 Crop Annotations & 45.31\% & 50.19\% & 45.74\% & 35.13\% \\
NB w/Top-5 Crop Annotations & 54.52\% & 65.82\% & 70.21\% & 75.67\%  \\
NB w/Top-20 Crop Annotations & 59.32\% & 69.66\% & 75.53\% & 78.37\% \\
ResNet & 58.02\% & 72.58\% & 76.01\% & 86.48\% \\

\hline
\end{tabular}

\caption{Example-level correlation accuracy between the intersection of two model tests and each one model test.}
\label{tab:dejavu_ref_example_test}

\end{table}

\subsection{Common memorized vs. correlated examples}\label{sec:common_mem_corr}
In this section we showcase examples of common dataset-level correlations and memorization by the OSS pre-trained representation models such as VICReg, Barlow Twins and Dino.

\begin{figure*}[tb]
    \centering
    \includegraphics[width=0.8\textwidth]{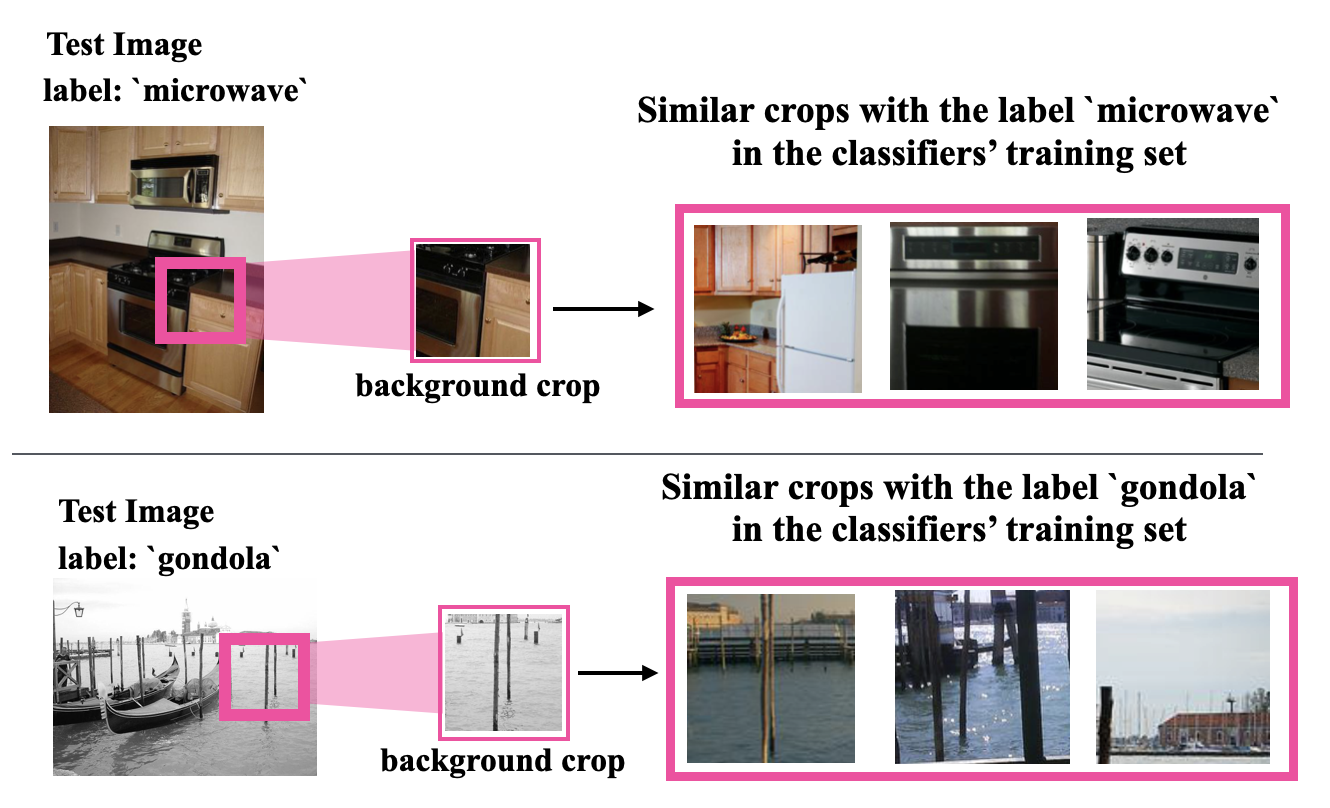}
    \caption{Two common dataset-level correlations: 1) `stove, kitchen` and `microwave`, 2) `sky, pole, water` and `gondola`. ResNet and Naive Bayes classifiers learn to associate the background crops with the image label.}
    \label{fig:corr_examples_oss}
\end{figure*}

\begin{figure*}[tb]
    \centering
    \includegraphics[width=0.8\textwidth]{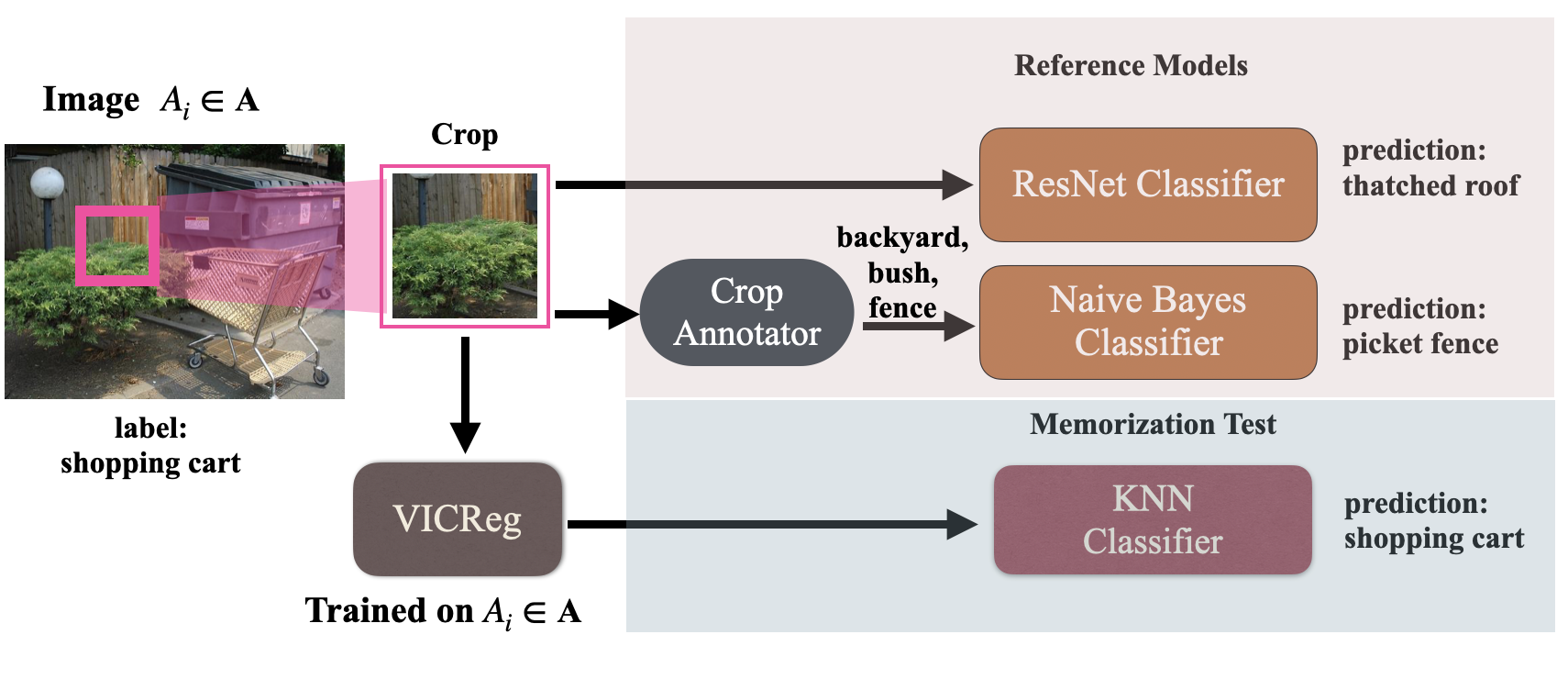}
    \caption{KNN predicts correct class `shopping cart` given VICReg's representation of the background crop. Here the original image of the crop is part of the VICReg's training. Resnet and NB classifiers, however, fail to predict the correct class which concludes that the image is memorized.}
    \label{fig:memorization_oob}
\end{figure*}

\begin{figure*}[tb]
    \centering
    \includegraphics[width=0.8\textwidth]{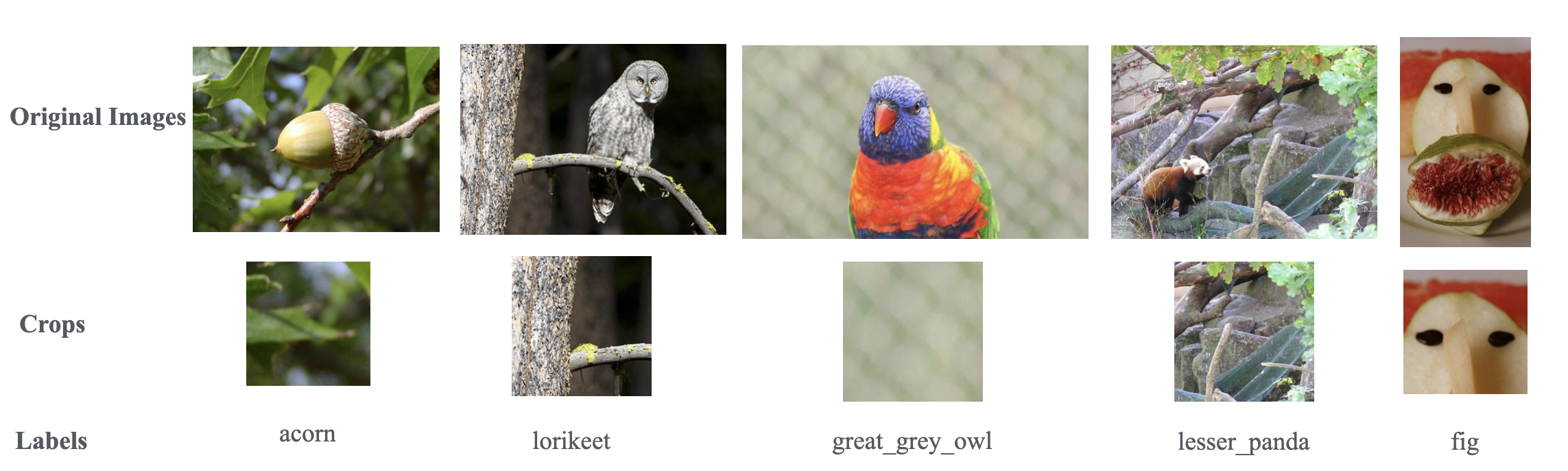}
    \caption{Top-5 images memorized by VICReg OSS model. Both Naive Bayes and ResNet classifiers fail to predict the correct class based on the background crops.}
    \label{fig:vision_top5_memorized}
\end{figure*}

\autoref{fig:corr_examples_oss} showcases examples of two common dataset-level correlations between `kitchen, store` and `microwave`, `gondola` and `pole, water`. Resnet and Naive Bayes classifiers learn these correlations effectively and help us distinguish memorization from dataset-level correlations.
In addition, we observe that memorization tends to happen in examples where there is no clear dataset-level correlations between the background crop and the foreground object. \autoref{fig:memorization_oob} demonstrates an example of a memorized image by VICReg pre-trained model. Here the reference models incorrectly predict the foreground object whereas KNN correctly classifies the VICReg representation of the crop. In this case, our approach identifies the image with the shopping cart as memorized. \autoref{fig:vision_top5_memorized} demonstrates top-5 images memorized by the VICReg OSS model. We observe that there is no clear correlation between the background and foreground and some images could be unique than the rest of the ImageNet.

\section{Additional Results for VLM One-Model D\'{e}j\`{a} Vu Memorization}\label{sec:vlm_additional_res}

\subsection{Experiment setup}
\label{sec:vlm_experiment_setup}

\paragraph{Dataset.} For the VLM experiments, we trained a CLIP model from scratch on the Shutterstock dataset---a private licensed data set consisting of 239M image–caption pairs. Since this dataset has many duplicate captions, we first perform caption-level de-duplication by considering only one image per caption. This resulted in 103M samples. We created three random splits of this data set of sizes 40M, 40M and 20M, we call these sets $D$, $\bar{D}$ and $D_\text{pub}$ respectively. We use the first two splits for training the CLIP models and use the $D_\text{pub}$ set for our nearest neighbor test to find the most relevant images to the target image from the training set $D$. 

\begin{figure*}[tb]
    \centering
    \includegraphics[width=\textwidth]{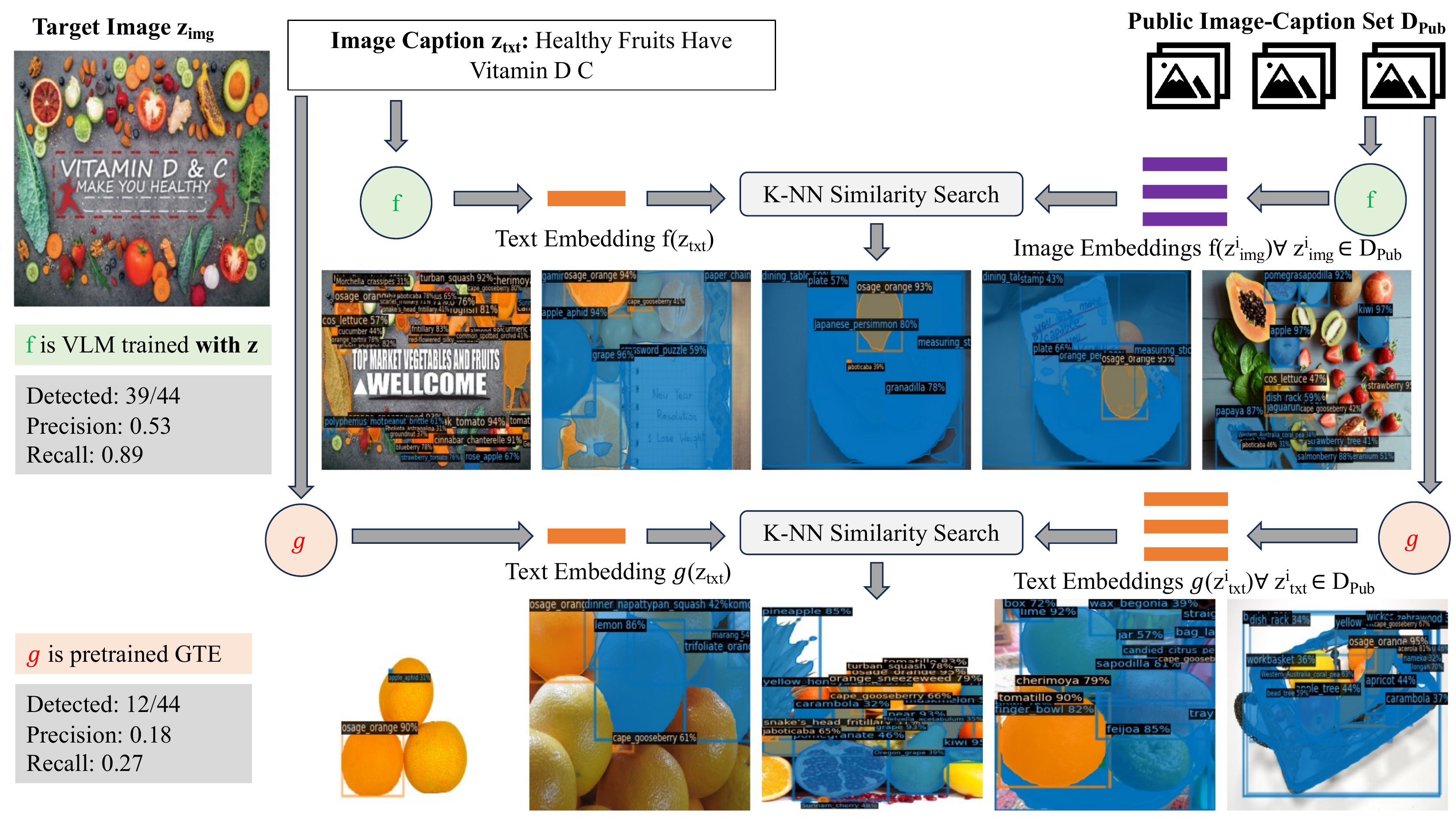}
    \caption{An example where an OpenCLIP model trained on a 40M subset of a Shutterstock data set exhibits \emph{d\'{e}j\`{a} vu memorization} of objects present in a training image. Public set is a separate collection of 20M images from Shutterstock that has no overlap with the training set. The objects annotated in {\color{orange} orange} are true positives, i.e., the ones present in the target image, and the objects annotated in {\color{blue} blue} are false positives. For the OpenCLIP model $f$ trained on the target image, our test recovers significantly more memorized objects compared to the pretrained GTE language model $g$ that finds the closest captions from the public set using the k-NN search in the text embedding space.}
    \label{fig:ss_example}
\end{figure*}

For object annotation on Shutterstock images, we use an open-source annotation tool, called Detic~\citep{zhou2022detecting}, that can annotate all the 21K ImageNet objects. We use a threshold of 0.3 to identify object bounding boxes (i.e., any bounding box that has more than 0.3 confidence is considered for annotation), as the default 0.5 threshold results in nearly 17\% images with no annotations. \autoref{fig:ss_example} shows the sample images with multiple object annotations obtained using Detic.

\paragraph{Model training.} The architecture we use is the ViT-B-32 model from OpenCLIP~\citep{openclip}. We train our models for 200 epochs with a learning rate of 0.0005 and a warmup of 2000 steps for cosine learning rate scheduler. Our training runs use 512GB RAM and use 32 Nvidia A100 GPUs with a global batch size of 32\,768. A single training run on 40M data size takes around 10 days. Our CLIP models trained on 40M data sets achieve around 41.26\% zero-shot classification accuracy on ImageNet data set, which is in line with CLIP models trained on similar size data sets.

\paragraph{Metrics.} To quantify \dejavu memorization for the target sample $z$, we consider the precision and recall metrics defined in \autoref{eq:prec_and_recall}. These metrics are scaled in a range $[0,1]$ and quantify memorization at an individual sample level. We can also quantify memorization at the dataset level using the population precision gap (PPG), population recall gap (PRG) and AUC gap (AUCG) metrics defined by \cite{jayaraman2024deja}. AUCG measures the gap between the cumulative object recall distributions of $f_A$ and $f_B$. The equations for PPG and PRG metrics are given below:
\begin{equation}
\begin{aligned}
    PPG = & \frac{1}{|A|}\Big(|\{z \in A : prec(z, f_A) > prec(z, f_B)\}| - |\{z \in A : prec(z, f_A) < prec(z, f_B)\}|\Big), \\
    PRG = & \frac{1}{|A|}\Big(|\{z \in A : rec(z, f_A) > rec(z, f_B)\}| - |\{z \in A : rec(z, f_A) < rec(z, f_B)\}|\Big),
\end{aligned}
\end{equation}

\subsection{Additional experiments}
\label{sec:vlm_additional_exp}

\begin{table}[ptb]
    \centering
    \begin{tabular}{|l|cc|cc|cc|}
        \hline
        Comparison & \multicolumn{2}{c|}{Using Top-1 NNs} & \multicolumn{2}{c|}{Using Top-10 NNs} & \multicolumn{2}{c|}{Using Top-100 NNs} \\ 
        & PPG & PRG & PPG & PRG & PPG & PRG \\ \hline
        Two Model & 0.030 & 0.030 & 0.060 & 0.064 & 0.054 & 0.063 \\
        $f_{t2i}$ vs $g$ & 0.014 & 0.034 & 0.094 & 0.082 & 0.190 & 0.107 \\
        $f_{t2t}$ vs $g$ & 0.033 & 0.038 & 0.098 & 0.065 & 0.218 & 0.061 \\ \hline 
    \end{tabular}
    \caption{Comparing the population-level memorization for predicting all objects for various settings where the 40M $D$ set is used as the target set. {\footnotesize For $g$, we use the GTE model where we match the target caption with the public set captions. For the VLMs, $t2i$ is the cross-modal setting where target caption is matched with public set images for kNN search, $t2t$ is the unimodal setting where only the text modality of the model is used for kNN search, i.e., target caption is matched with public set captions similar to the $g$ case. We do not consider the image-to-image search as the target image is not known to the adversary.}}
    \label{tab:vlm_dejavu_40M_nns}
\end{table}

\begin{table}[ptb]
    \centering
    \begin{tabular}{|l|cc|cc|cc|}
        \hline
        Comparison & \multicolumn{2}{c|}{Predicting Top-1 Object} & \multicolumn{2}{c|}{Predicting Top-10 Objects} & \multicolumn{2}{c|}{Predicting All Objects} \\ 
        & PPG & PRG & PPG & PRG & PPG & PRG \\ \hline
        Two Model & 0.022 & 0.022 & 0.055 & 0.056 & 0.060 & 0.064 \\
        $f_{t2i}$ vs $g$ & 0.034 & 0.034 & 0.081 & 0.074 & 0.094 & 0.082 \\
        $f_{t2t}$ vs $g$ & 0.026 & 0.026 & 0.068 & 0.064 & 0.098 & 0.065 \\ \hline
    \end{tabular}
    \caption{Comparing the population-level memorization for various settings with top-10 public NNs where the 40M $D$ set is used as the target set. {\footnotesize For $g$, we use the GTE model where we match the target caption with the public set captions. For the VLMs, $t2i$ is the cross-modal setting where target caption is matched with public set images for kNN search, $t2t$ is the unimodal setting where only the text modality of the model is used for kNN search, i.e., target caption is matched with public set captions similar to the $g$ case. We do not consider the image-to-image search as the target image is not known to the adversary.}}
    \label{tab:vlm_dejavu_40M_labels}
\end{table}

\begin{table}[ptb]
    \centering
    \begin{tabular}{|l|cc|cc|cc|}
        \hline
        Comparison & \multicolumn{2}{c|}{Using Top-1 NNs} & \multicolumn{2}{c|}{Using Top-10 NNs} & \multicolumn{2}{c|}{Using Top-100 NNs} \\ 
        & PPG & PRG & PPG & PRG & PPG & PRG \\ \hline
        $f_{t2i}$ vs $g$ & 0.140 & 0.164 & 0.156 & 0.165 & 0.257 & 0.097 \\
        $f_{t2t}$ vs $g$ & 0.053 & 0.061 & 0.054 & 0.092 & 0.066 & 0.080 \\ \hline 
    \end{tabular}
    \caption{Population-level memorization for predicting all objects with the pre-trained YFCC15M OSS model. {\footnotesize For $g$, we use the GTE model where we match the target caption with the public set captions. For the VLMs, $t2i$ is the cross-modal setting where target caption is matched with public set images for kNN search, $t2t$ is the unimodal setting where only the text modality of the model is used for kNN search, i.e., target caption is matched with public set captions similar to the $g$ case. We do not consider the image-to-image search as the target image is not known to the adversary.}}
    \label{tab:vlm_oob_yfcc15m_nns}
\end{table}

\begin{table}[ptb]
    \centering
    \begin{tabular}{|l|cc|cc|cc|}
        \hline
        Comparison & \multicolumn{2}{c|}{Predicting Top-1 Object} & \multicolumn{2}{c|}{Predicting Top-10 Objects} & \multicolumn{2}{c|}{Predicting All Objects} \\ 
        & PPG & PRG & PPG & PRG & PPG & PRG \\ \hline
        $f_{t2i}$ vs $g$ & 0.092 & 0.092 & 0.143 & 0.142 & 0.156 & 0.165 \\
        $f_{t2t}$ vs $g$ & 0.054 & 0.054 & 0.071 & 0.071 & 0.054 & 0.092 \\ \hline
    \end{tabular}
    \caption{Population-level memorization for predicting top-$k$ objects with top-10 public NNs with the pre-trained YFCC15M OSS model. {\footnotesize For $g$, we use the GTE model where we match the target caption with the public set captions. For the VLMs, $t2i$ is the cross-modal setting where target caption is matched with public set images for kNN search, $t2t$ is the unimodal setting where only the text modality of the model is used for kNN search, i.e., target caption is matched with public set captions similar to the $g$ case. We do not consider the image-to-image search as the target image is not known to the adversary.}}
    \label{tab:vlm_oob_yfcc15m_labels}
\end{table}

\begin{figure}[tb]
    \centering
    \begin{subfigure}{\textwidth}
    \centering
    \includegraphics[width=0.85\linewidth]{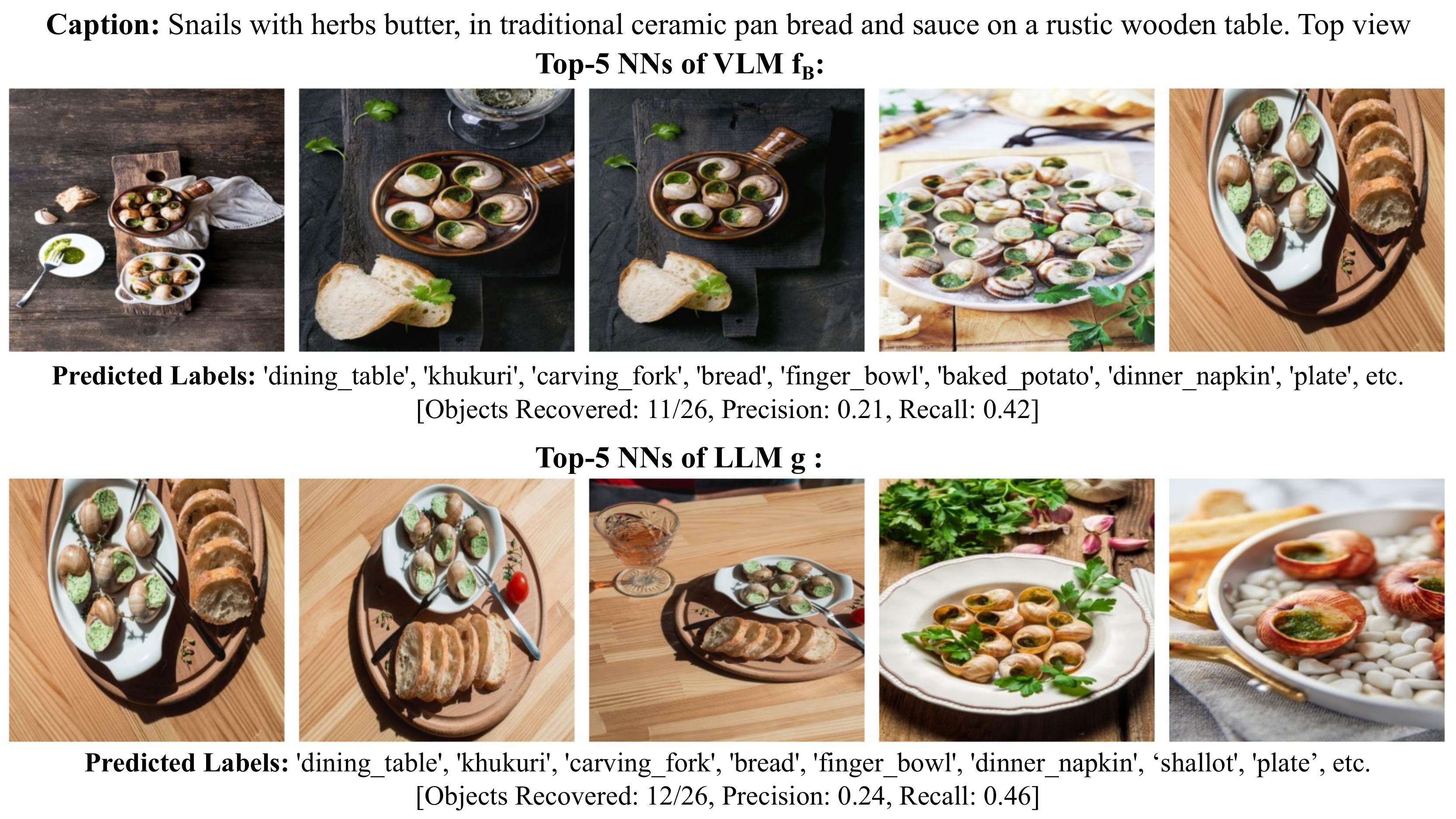}
    \caption{Example where both $f_B$ and $g$ perform similarly.}
    \end{subfigure}
    \begin{subfigure}{\textwidth}
    \centering
    \includegraphics[width=0.85\linewidth]{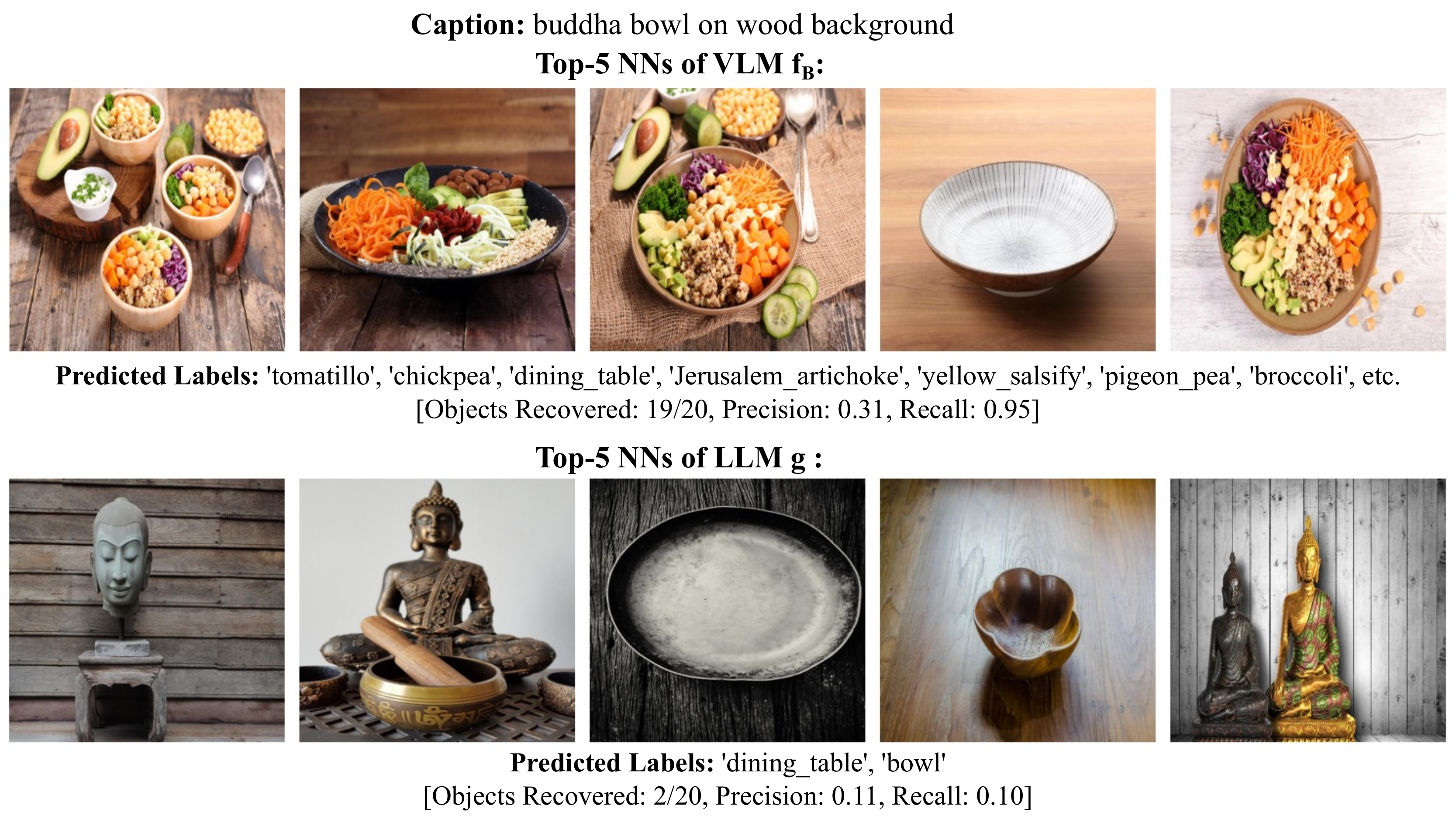}
    \caption{Example where $f_B$ performs better than $g$.}
    \end{subfigure}
    \begin{subfigure}{\textwidth}
    \centering
    \includegraphics[width=0.85\linewidth]{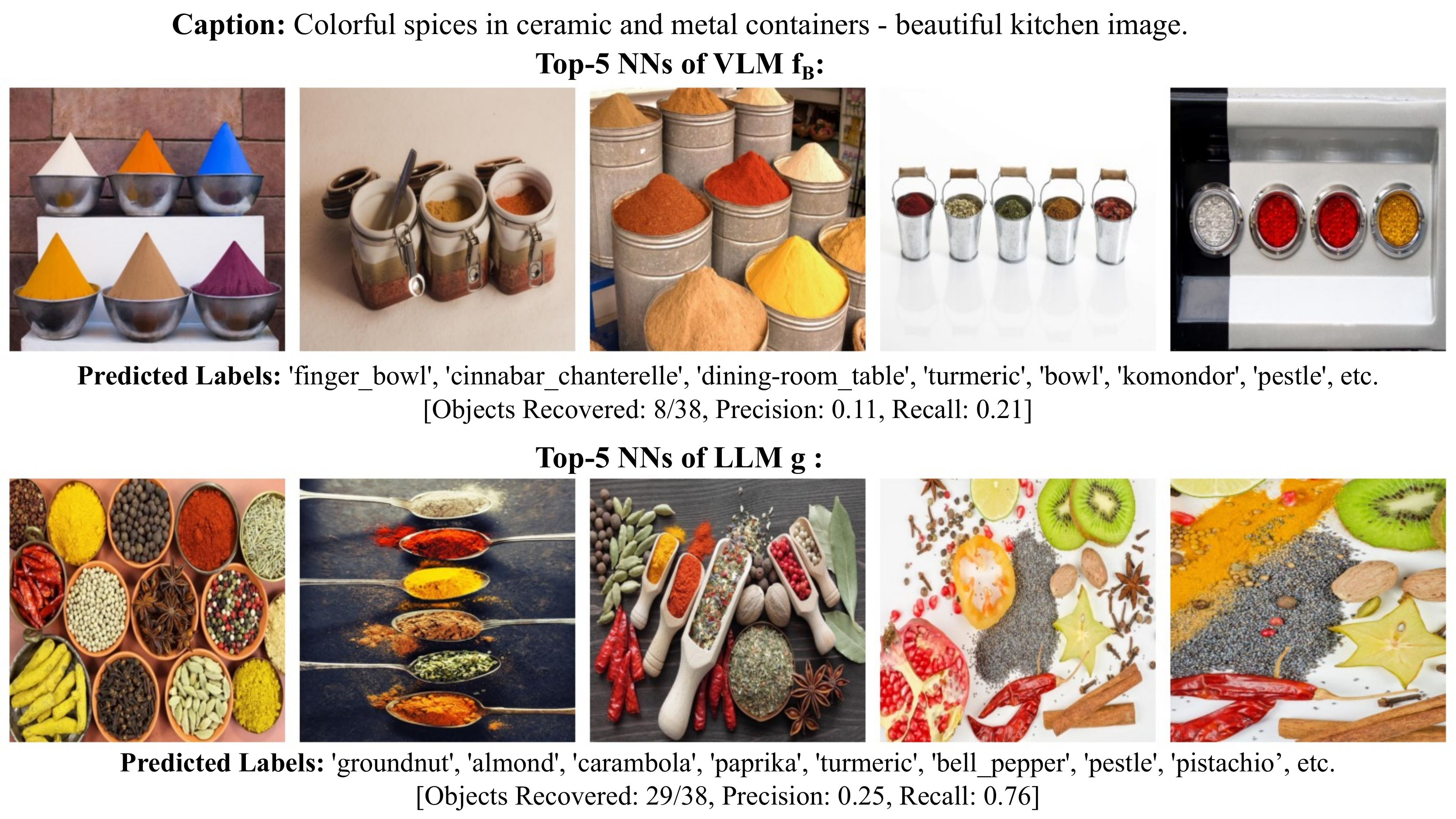}
    \caption{Example where $g$ performs better than $f_B$.}
    \end{subfigure}
    \caption{Examples showing correlations captured by the reference VLM ($f_B$) and the LLM ($g$).}
    \label{fig:vlm_corr_examples}
\end{figure}

\begin{figure}[tb]
    \centering
    \begin{subfigure}{\textwidth}
    \includegraphics[width=\linewidth]{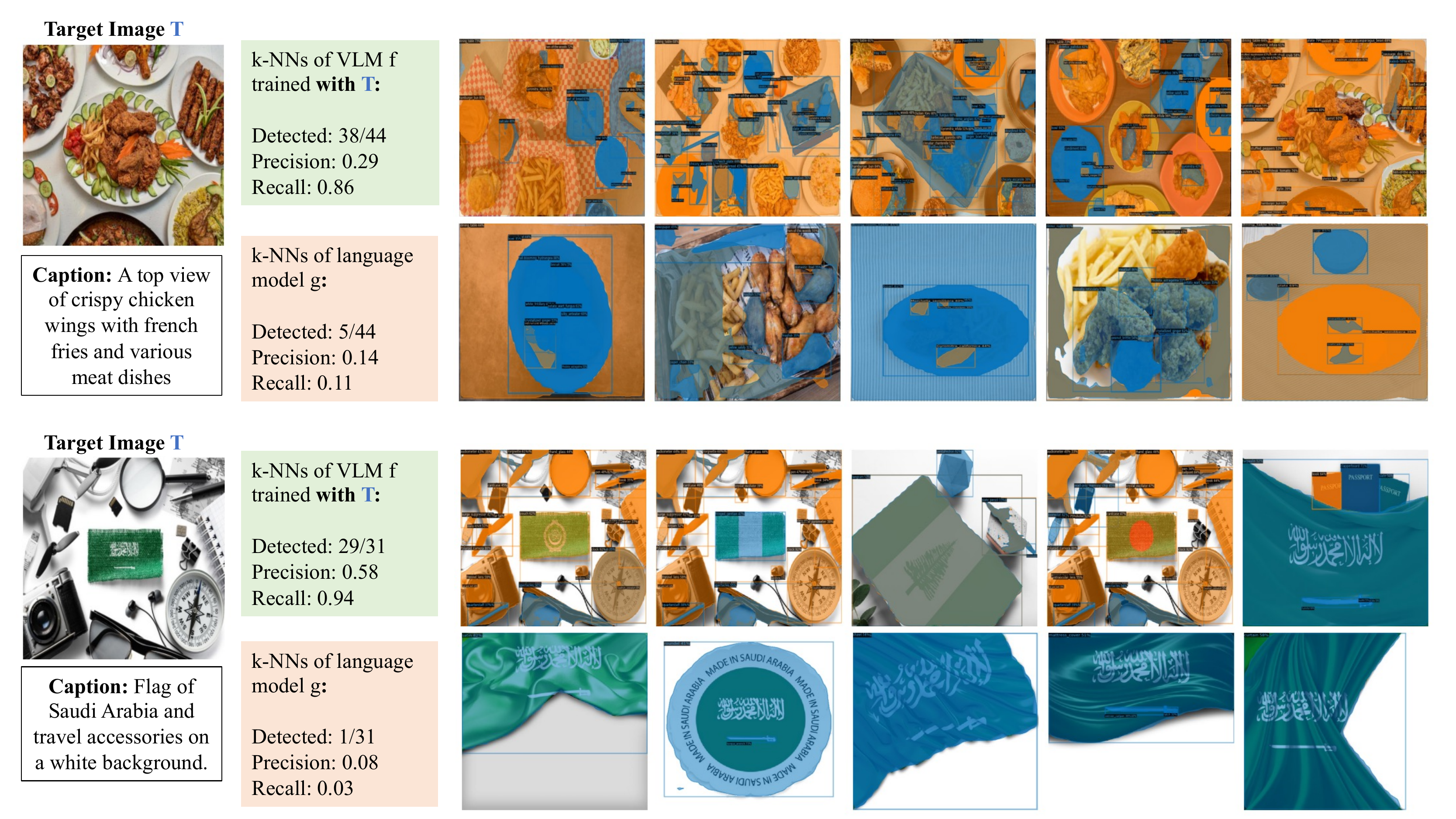}
    \end{subfigure}
    \begin{subfigure}{\textwidth}
    \includegraphics[width=\linewidth]{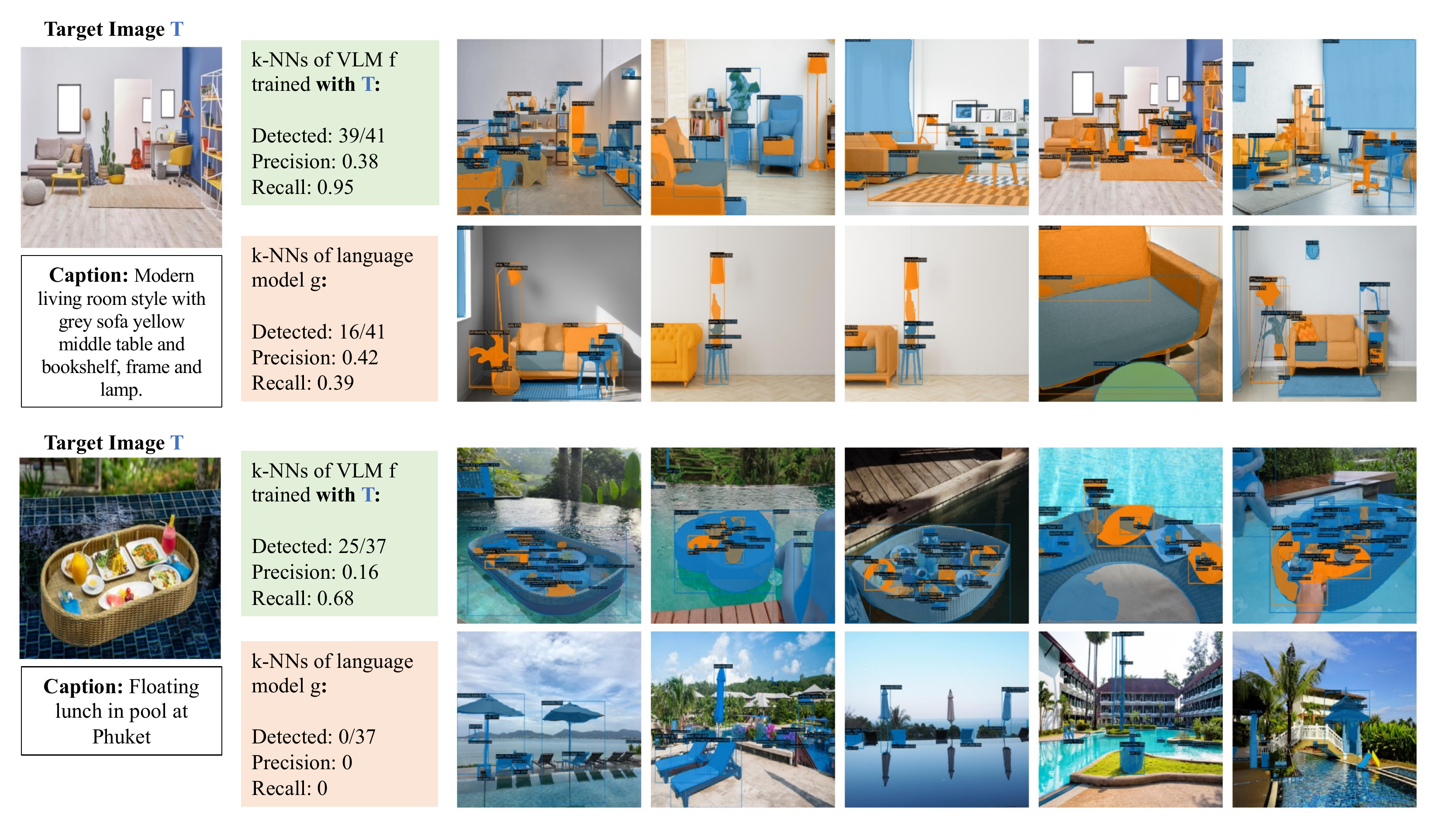}
    \end{subfigure}
    \caption{Additional qualitative examples from Shutterstock data set showing examples memorized by the VLM.}
    \label{fig:vlm_memorized_examples}
\end{figure}

\clearpage
\newpage
\section*{NeurIPS Paper Checklist}

The checklist is designed to encourage best practices for responsible machine learning research, addressing issues of reproducibility, transparency, research ethics, and societal impact. Do not remove the checklist: {\bf The papers not including the checklist will be desk rejected.} The checklist should follow the references and follow the (optional) supplemental material.  The checklist does NOT count towards the page
limit. 

Please read the checklist guidelines carefully for information on how to answer these questions. For each question in the checklist:
\begin{itemize}
    \item You should answer \answerYes{}, \answerNo{}, or \answerNA{}.
    \item \answerNA{} means either that the question is Not Applicable for that particular paper or the relevant information is Not Available.
    \item Please provide a short (1–2 sentence) justification right after your answer (even for NA). 
\end{itemize}

{\bf The checklist answers are an integral part of your paper submission.} They are visible to the reviewers, area chairs, senior area chairs, and ethics reviewers. You will be asked to also include it (after eventual revisions) with the final version of your paper, and its final version will be published with the paper.

The reviewers of your paper will be asked to use the checklist as one of the factors in their evaluation. While "\answerYes{}" is generally preferable to "\answerNo{}", it is perfectly acceptable to answer "\answerNo{}" provided a proper justification is given (e.g., "error bars are not reported because it would be too computationally expensive" or "we were unable to find the license for the dataset we used"). In general, answering "\answerNo{}" or "\answerNA{}" is not grounds for rejection. While the questions are phrased in a binary way, we acknowledge that the true answer is often more nuanced, so please just use your best judgment and write a justification to elaborate. All supporting evidence can appear either in the main paper or the supplemental material, provided in appendix. If you answer \answerYes{} to a question, in the justification please point to the section(s) where related material for the question can be found.

IMPORTANT, please:
\begin{itemize}
    \item {\bf Delete this instruction block, but keep the section heading ``NeurIPS paper checklist"},
    \item  {\bf Keep the checklist subsection headings, questions/answers and guidelines below.}
    \item {\bf Do not modify the questions and only use the provided macros for your answers}.
\end{itemize}


\begin{enumerate}

\item {\bf Claims}
    \item[] Question: Do the main claims made in the abstract and introduction accurately reflect the paper's contributions and scope?
    \item[] Answer: \answerYes{} 
    \item[] Justification: Section ~\ref{sec:dataset_level_corr} formally describes our methods to identify dataset-level correlations. It showcases the effectiveness of our method for vision (VICReg, Barlow Twins and DINO) and vision language (CLIP) models. Section ~\ref{sec:vision_vlm_mem} discusses the memorization in both OSS models and models trained on subsets of the training data. The experimental results show that our method effectively identifies memorization in those models.
    \item[] Guidelines:
    \begin{itemize}
        \item The answer NA means that the abstract and introduction do not include the claims made in the paper.
        \item The abstract and/or introduction should clearly state the claims made, including the contributions made in the paper and important assumptions and limitations. A No or NA answer to this question will not be perceived well by the reviewers. 
        \item The claims made should match theoretical and experimental results, and reflect how much the results can be expected to generalize to other settings. 
        \item It is fine to include aspirational goals as motivation as long as it is clear that these goals are not attained by the paper. 
    \end{itemize}

\item {\bf Limitations}
    \item[] Question: Does the paper discuss the limitations of the work performed by the authors?
    \item[] Answer: \answerYes{} 
    \item[] Justification: We present the limitations of our methods at the end of section 3 and 4.
    \item[] Guidelines:
    \begin{itemize}
        \item The answer NA means that the paper has no limitation while the answer No means that the paper has limitations, but those are not discussed in the paper. 
        \item The authors are encouraged to create a separate "Limitations" section in their paper.
        \item The paper should point out any strong assumptions and how robust the results are to violations of these assumptions (e.g., independence assumptions, noiseless settings, model well-specification, asymptotic approximations only holding locally). The authors should reflect on how these assumptions might be violated in practice and what the implications would be.
        \item The authors should reflect on the scope of the claims made, e.g., if the approach was only tested on a few datasets or with a few runs. In general, empirical results often depend on implicit assumptions, which should be articulated.
        \item The authors should reflect on the factors that influence the performance of the approach. For example, a facial recognition algorithm may perform poorly when image resolution is low or images are taken in low lighting. Or a speech-to-text system might not be used reliably to provide closed captions for online lectures because it fails to handle technical jargon.
        \item The authors should discuss the computational efficiency of the proposed algorithms and how they scale with dataset size.
        \item If applicable, the authors should discuss possible limitations of their approach to address problems of privacy and fairness.
        \item While the authors might fear that complete honesty about limitations might be used by reviewers as grounds for rejection, a worse outcome might be that reviewers discover limitations that aren't acknowledged in the paper. The authors should use their best judgment and recognize that individual actions in favor of transparency play an important role in developing norms that preserve the integrity of the community. Reviewers will be specifically instructed to not penalize honesty concerning limitations.
    \end{itemize}

\item {\bf Theory Assumptions and Proofs}
    \item[] Question: For each theoretical result, does the paper provide the full set of assumptions and a complete (and correct) proof?
    \item[] Answer: \answerNA{} 
    \item[] Justification: 
    
    \item[] Guidelines:
    \begin{itemize}
        \item The answer NA means that the paper does not include theoretical results. 
        \item All the theorems, formulas, and proofs in the paper should be numbered and cross-referenced.
        \item All assumptions should be clearly stated or referenced in the statement of any theorems.
        \item The proofs can either appear in the main paper or the supplemental material, but if they appear in the supplemental material, the authors are encouraged to provide a short proof sketch to provide intuition. 
        \item Inversely, any informal proof provided in the core of the paper should be complemented by formal proofs provided in appendix or supplemental material.
        \item Theorems and Lemmas that the proof relies upon should be properly referenced. 
    \end{itemize}

    \item {\bf Experimental Result Reproducibility}
    \item[] Question: Does the paper fully disclose all the information needed to reproduce the main experimental results of the paper to the extent that it affects the main claims and/or conclusions of the paper (regardless of whether the code and data are provided or not)?
    \item[] Answer: \answerYes{} 
    \item[] Justification: We put the experimental details in the appendix.
    \item[] Guidelines:
    \begin{itemize}
        \item The answer NA means that the paper does not include experiments.
        \item If the paper includes experiments, a No answer to this question will not be perceived well by the reviewers: Making the paper reproducible is important, regardless of whether the code and data are provided or not.
        \item If the contribution is a dataset and/or model, the authors should describe the steps taken to make their results reproducible or verifiable. 
        \item Depending on the contribution, reproducibility can be accomplished in various ways. For example, if the contribution is a novel architecture, describing the architecture fully might suffice, or if the contribution is a specific model and empirical evaluation, it may be necessary to either make it possible for others to replicate the model with the same dataset, or provide access to the model. In general. releasing code and data is often one good way to accomplish this, but reproducibility can also be provided via detailed instructions for how to replicate the results, access to a hosted model (e.g., in the case of a large language model), releasing of a model checkpoint, or other means that are appropriate to the research performed.
        \item While NeurIPS does not require releasing code, the conference does require all submissions to provide some reasonable avenue for reproducibility, which may depend on the nature of the contribution. For example
        \begin{enumerate}
            \item If the contribution is primarily a new algorithm, the paper should make it clear how to reproduce that algorithm.
            \item If the contribution is primarily a new model architecture, the paper should describe the architecture clearly and fully.
            \item If the contribution is a new model (e.g., a large language model), then there should either be a way to access this model for reproducing the results or a way to reproduce the model (e.g., with an open-source dataset or instructions for how to construct the dataset).
            \item We recognize that reproducibility may be tricky in some cases, in which case authors are welcome to describe the particular way they provide for reproducibility. In the case of closed-source models, it may be that access to the model is limited in some way (e.g., to registered users), but it should be possible for other researchers to have some path to reproducing or verifying the results.
        \end{enumerate}
    \end{itemize}

\item {\bf Open access to data and code}
    \item[] Question: Does the paper provide open access to the data and code, with sufficient instructions to faithfully reproduce the main experimental results, as described in supplemental material?
    \item[] Answer: \answerNo{} 
    \item[] Justification: The code is released on github. The links are provided in the abstract.
    \item[] Guidelines:
    \begin{itemize}
        \item The answer NA means that paper does not include experiments requiring code.
        \item Please see the NeurIPS code and data submission guidelines (\url{https://nips.cc/public/guides/CodeSubmissionPolicy}) for more details.
        \item While we encourage the release of code and data, we understand that this might not be possible, so “No” is an acceptable answer. Papers cannot be rejected simply for not including code, unless this is central to the contribution (e.g., for a new open-source benchmark).
        \item The instructions should contain the exact command and environment needed to run to reproduce the results. See the NeurIPS code and data submission guidelines (\url{https://nips.cc/public/guides/CodeSubmissionPolicy}) for more details.
        \item The authors should provide instructions on data access and preparation, including how to access the raw data, preprocessed data, intermediate data, and generated data, etc.
        \item The authors should provide scripts to reproduce all experimental results for the new proposed method and baselines. If only a subset of experiments are reproducible, they should state which ones are omitted from the script and why.
        \item At submission time, to preserve anonymity, the authors should release anonymized versions (if applicable).
        \item Providing as much information as possible in supplemental material (appended to the paper) is recommended, but including URLs to data and code is permitted.
    \end{itemize}

\item {\bf Experimental Setting/Details}
    \item[] Question: Does the paper specify all the training and test details (e.g., data splits, hyperparameters, how they were chosen, type of optimizer, etc.) necessary to understand the results?
    \item[] Answer: \answerYes{} 
    \item[] Justification: We specify all the training and test details in the appendix.
    \item[] Guidelines:
    \begin{itemize}
        \item The answer NA means that the paper does not include experiments.
        \item The experimental setting should be presented in the core of the paper to a level of detail that is necessary to appreciate the results and make sense of them.
        \item The full details can be provided either with the code, in appendix, or as supplemental material.
    \end{itemize}

\item {\bf Experiment Statistical Significance}
    \item[] Question: Does the paper report error bars suitably and correctly defined or other appropriate information about the statistical significance of the experiments?
    \item[] Answer: \answerNo{} 
    \item[] Justification: We believe that by using different models, we have already a good estimate of how much noise there can be in our process.
    \item[] Guidelines:
    \begin{itemize}
        \item The answer NA means that the paper does not include experiments.
        \item The authors should answer "Yes" if the results are accompanied by error bars, confidence intervals, or statistical significance tests, at least for the experiments that support the main claims of the paper.
        \item The factors of variability that the error bars are capturing should be clearly stated (for example, train/test split, initialization, random drawing of some parameter, or overall run with given experimental conditions).
        \item The method for calculating the error bars should be explained (closed form formula, call to a library function, bootstrap, etc.)
        \item The assumptions made should be given (e.g., Normally distributed errors).
        \item It should be clear whether the error bar is the standard deviation or the standard error of the mean.
        \item It is OK to report 1-sigma error bars, but one should state it. The authors should preferably report a 2-sigma error bar than state that they have a 96\% CI, if the hypothesis of Normality of errors is not verified.
        \item For asymmetric distributions, the authors should be careful not to show in tables or figures symmetric error bars that would yield results that are out of range (e.g. negative error rates).
        \item If error bars are reported in tables or plots, The authors should explain in the text how they were calculated and reference the corresponding figures or tables in the text.
    \end{itemize}

\item {\bf Experiments Compute Resources}
    \item[] Question: For each experiment, does the paper provide sufficient information on the computer resources (type of compute workers, memory, time of execution) needed to reproduce the experiments?
    \item[] Answer: \answerYes{} 
    \item[] Justification: Compute resources are described in sections ~\ref{section: vision_exp} and ~\ref{sec:vlm_experiment_setup}.
    \item[] Guidelines:
    \begin{itemize}
        \item The answer NA means that the paper does not include experiments.
        \item The paper should indicate the type of compute workers CPU or GPU, internal cluster, or cloud provider, including relevant memory and storage.
        \item The paper should provide the amount of compute required for each of the individual experimental runs as well as estimate the total compute. 
        \item The paper should disclose whether the full research project required more compute than the experiments reported in the paper (e.g., preliminary or failed experiments that didn't make it into the paper). 
    \end{itemize}
    
\item {\bf Code Of Ethics}
    \item[] Question: Does the research conducted in the paper conform, in every respect, with the NeurIPS Code of Ethics \url{https://neurips.cc/public/EthicsGuidelines}?
    \item[] Answer: \answerYes{} 
    \item[] Justification: The paper conforms with the NeurIPS Code of Ethics in every respect.
    \item[] Guidelines:
    \begin{itemize}
        \item The answer NA means that the authors have not reviewed the NeurIPS Code of Ethics.
        \item If the authors answer No, they should explain the special circumstances that require a deviation from the Code of Ethics.
        \item The authors should make sure to preserve anonymity (e.g., if there is a special consideration due to laws or regulations in their jurisdiction).
    \end{itemize}

\item {\bf Broader Impacts}
    \item[] Question: Does the paper discuss both potential positive societal impacts and negative societal impacts of the work performed?
    \item[] Answer: \answerYes{} 
    \item[] Justification: Our work has potential to lead to better tools for measuring privacy leakage stemming from memorization.
    \item[] Guidelines:
    \begin{itemize}
        \item The answer NA means that there is no societal impact of the work performed.
        \item If the authors answer NA or No, they should explain why their work has no societal impact or why the paper does not address societal impact.
        \item Examples of negative societal impacts include potential malicious or unintended uses (e.g., disinformation, generating fake profiles, surveillance), fairness considerations (e.g., deployment of technologies that could make decisions that unfairly impact specific groups), privacy considerations, and security considerations.
        \item The conference expects that many papers will be foundational research and not tied to particular applications, let alone deployments. However, if there is a direct path to any negative applications, the authors should point it out. For example, it is legitimate to point out that an improvement in the quality of generative models could be used to generate deepfakes for disinformation. On the other hand, it is not needed to point out that a generic algorithm for optimizing neural networks could enable people to train models that generate Deepfakes faster.
        \item The authors should consider possible harms that could arise when the technology is being used as intended and functioning correctly, harms that could arise when the technology is being used as intended but gives incorrect results, and harms following from (intentional or unintentional) misuse of the technology.
        \item If there are negative societal impacts, the authors could also discuss possible mitigation strategies (e.g., gated release of models, providing defenses in addition to attacks, mechanisms for monitoring misuse, mechanisms to monitor how a system learns from feedback over time, improving the efficiency and accessibility of ML).
    \end{itemize}
    
\item {\bf Safeguards}
    \item[] Question: Does the paper describe safeguards that have been put in place for responsible release of data or models that have a high risk for misuse (e.g., pretrained language models, image generators, or scraped datasets)?
    \item[] Answer: \answerNA{} 
    \item[] Justification: 
    \item[] Guidelines:
    \begin{itemize}
        \item The answer NA means that the paper poses no such risks.
        \item Released models that have a high risk for misuse or dual-use should be released with necessary safeguards to allow for controlled use of the model, for example by requiring that users adhere to usage guidelines or restrictions to access the model or implementing safety filters. 
        \item Datasets that have been scraped from the Internet could pose safety risks. The authors should describe how they avoided releasing unsafe images.
        \item We recognize that providing effective safeguards is challenging, and many papers do not require this, but we encourage authors to take this into account and make a best faith effort.
    \end{itemize}

\item {\bf Licenses for existing assets}
    \item[] Question: Are the creators or original owners of assets (e.g., code, data, models), used in the paper, properly credited and are the license and terms of use explicitly mentioned and properly respected?
    \item[] Answer: \answerYes{} 
    \item[] Justification: We cite the licenses in the appendix \ref{sec:licenses}.
    \item[] Guidelines:
    \begin{itemize}
        \item The answer NA means that the paper does not use existing assets.
        \item The authors should cite the original paper that produced the code package or dataset.
        \item The authors should state which version of the asset is used and, if possible, include a URL.
        \item The name of the license (e.g., CC-BY 4.0) should be included for each asset.
        \item For scraped data from a particular source (e.g., website), the copyright and terms of service of that source should be provided.
        \item If assets are released, the license, copyright information, and terms of use in the package should be provided. For popular datasets, \url{paperswithcode.com/datasets} has curated licenses for some datasets. Their licensing guide can help determine the license of a dataset.
        \item For existing datasets that are re-packaged, both the original license and the license of the derived asset (if it has changed) should be provided.
        \item If this information is not available online, the authors are encouraged to reach out to the asset's creators.
    \end{itemize}

\item {\bf New Assets}
    \item[] Question: Are new assets introduced in the paper well documented and is the documentation provided alongside the assets?
    \item[] Answer: \answerNA{} 
    \item[] Justification: 
    \item[] Guidelines:
    \begin{itemize}
        \item The answer NA means that the paper does not release new assets.
        \item Researchers should communicate the details of the dataset/code/model as part of their submissions via structured templates. This includes details about training, license, limitations, etc. 
        \item The paper should discuss whether and how consent was obtained from people whose asset is used.
        \item At submission time, remember to anonymize your assets (if applicable). You can either create an anonymized URL or include an anonymized zip file.
    \end{itemize}

\item {\bf Crowdsourcing and Research with Human Subjects}
    \item[] Question: For crowdsourcing experiments and research with human subjects, does the paper include the full text of instructions given to participants and screenshots, if applicable, as well as details about compensation (if any)? 
    \item[] Answer: \answerNA{} 
    \item[] Justification: 
    \item[] Guidelines:
    \begin{itemize}
        \item The answer NA means that the paper does not involve crowdsourcing nor research with human subjects.
        \item Including this information in the supplemental material is fine, but if the main contribution of the paper involves human subjects, then as much detail as possible should be included in the main paper. 
        \item According to the NeurIPS Code of Ethics, workers involved in data collection, curation, or other labor should be paid at least the minimum wage in the country of the data collector. 
    \end{itemize}

\item {\bf Institutional Review Board (IRB) Approvals or Equivalent for Research with Human Subjects}
    \item[] Question: Does the paper describe potential risks incurred by study participants, whether such risks were disclosed to the subjects, and whether Institutional Review Board (IRB) approvals (or an equivalent approval/review based on the requirements of your country or institution) were obtained?
    \item[] Answer: \answerNA{} 
    \item[] Justification: 
    \item[] Guidelines:
    \begin{itemize}
        \item The answer NA means that the paper does not involve crowdsourcing nor research with human subjects.
        \item Depending on the country in which research is conducted, IRB approval (or equivalent) may be required for any human subjects research. If you obtained IRB approval, you should clearly state this in the paper. 
        \item We recognize that the procedures for this may vary significantly between institutions and locations, and we expect authors to adhere to the NeurIPS Code of Ethics and the guidelines for their institution. 
        \item For initial submissions, do not include any information that would break anonymity (if applicable), such as the institution conducting the review.
    \end{itemize}

\end{enumerate}

\end{document}